\documentclass[10pt,twocolumn,letterpaper]{article}

\usepackage[pagenumbers]{cvpr} % To force page numbers, e.g. for an arXiv version

\usepackage{graphicx}
\usepackage{amsmath}
\usepackage{amssymb}
\usepackage{booktabs}
\usepackage{ragged2e} 
\usepackage{booktabs,makecell, multirow, tabularx}
\usepackage{appendix}
\usepackage{cite}
\usepackage{multicol}
\usepackage[pagebackref,breaklinks,colorlinks]{hyperref}

\usepackage[capitalize]{cleveref}
\crefname{section}{Sec.}{Secs.}
\Crefname{section}{Section}{Sections}
\Crefname{table}{Table}{Tables}
\crefname{table}{Tab.}{Tabs.}

\def\cvprPaperID{9131} % *** Enter the CVPR Paper ID here
\def\confName{ICCV}
\def\confYear{2023}

\begin{document}

\title{Multi-scale Alternated Attention Transformer for Generalized Stereo Matching}

\author{Wei Miao\\
\and
Hong Zhao\\
\and
Tongjia Chen\\
\and
Wei Huang\\
\and
Changyan Xiao\\
\and
College of Electrical and Information Engineering, Hunan University, Changsha, China\\
}
\maketitle

%%%%%%%%% ABSTRACT
\begin{abstract}
   Recent stereo matching networks achieves dramatic performance by introducing epipolar line constraint to limit the matching range of dual-view. However, in complicated real-world scenarios, the feature information based on intra-epipolar line alone is too weak to facilitate stereo matching. In this paper, we present a simple but highly effective network called Alternated Attention U-shaped Transformer (AAUformer) to balance the impact of epipolar line in dual and single view respectively for excellent generalization performance. Compared to other models, our model has several main designs: 1) to better liberate the local semantic features of the single-view at pixel level, we introduce window self-attention to break the limits of intra-row self-attention and completely replace the convolutional network for denser features before cross-matching; 2) the multi-scale alternated attention backbone network was designed to extract invariant features in order to achieves the coarse-to-fine matching process for hard-to-discriminate regions. We performed a series of both comparative studies and ablation studies on several mainstream stereo matching datasets. The results demonstrate that our model achieves state-of-the-art on the Scene Flow dataset, and the fine-tuning performance is competitive on the KITTI 2015 dataset. In addition, for cross generalization experiments on synthetic and real-world datasets, our model outperforms several state-of-the-art works.
\end{abstract}
\begin{figure}[htbp]
    \centering
	\begin{subfigure}{0.96\linewidth}
		\begin{subfigure}{0.32\linewidth}
		  \centering
		  \includegraphics[width=0.96\linewidth]{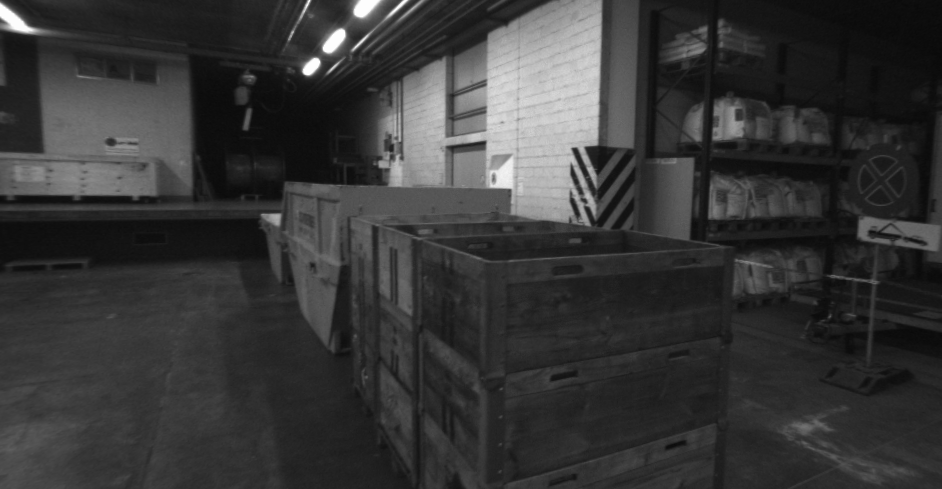}
	  \end{subfigure}
        \begin{subfigure}{0.32\linewidth}
		  \centering
		  \includegraphics[width=0.96\linewidth]{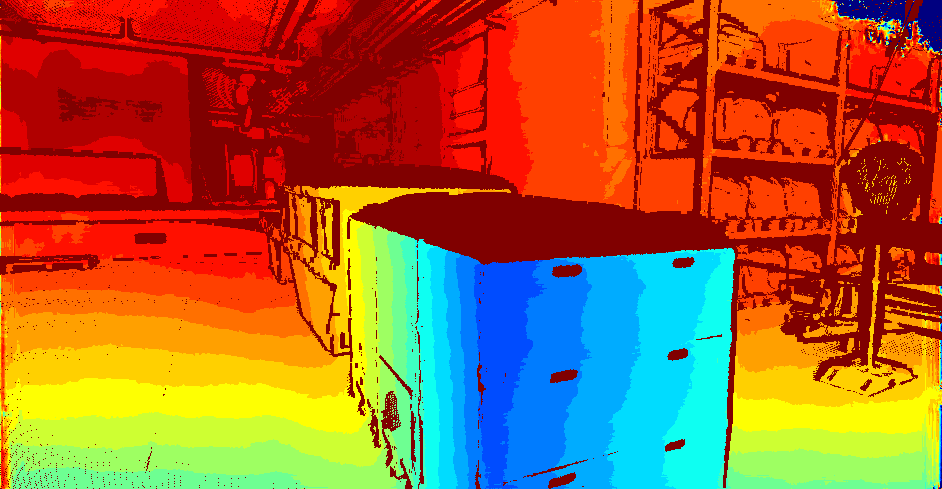}
	  \end{subfigure}
        \begin{subfigure}{0.32\linewidth}
		  \centering
		  \includegraphics[width=0.96\linewidth]{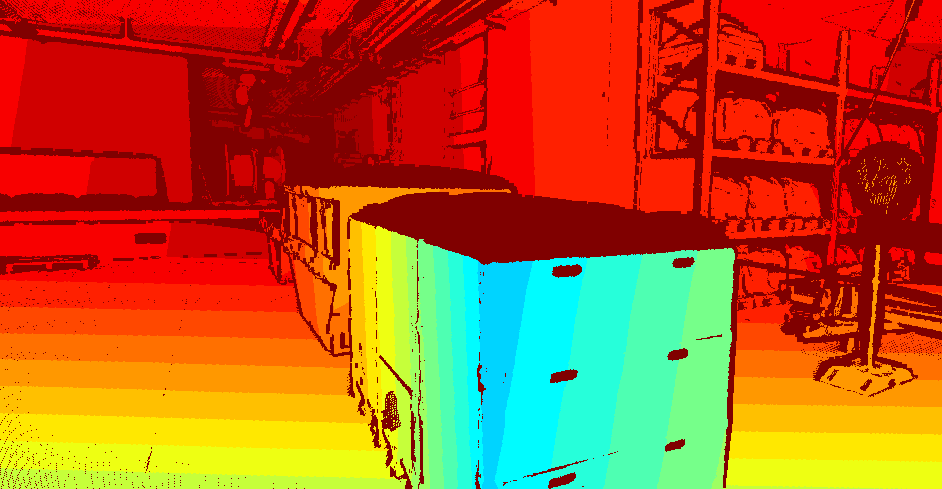}
	  \end{subfigure}
        \label{1-a}%文中引用该图片代号
	\end{subfigure}
    \centering
    \begin{subfigure}{0.96\linewidth}
		\begin{subfigure}{0.32\linewidth}
		  \centering
		  \includegraphics[width=0.96\linewidth]{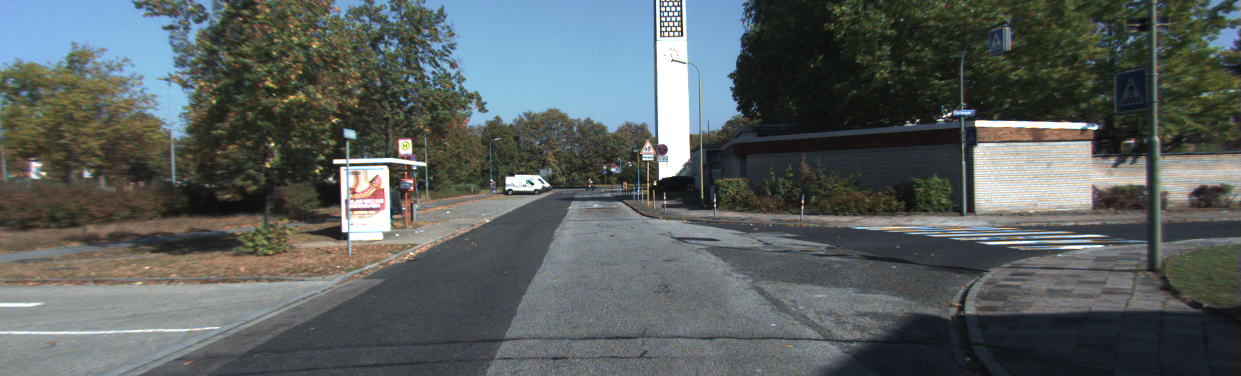}
	  \end{subfigure}
        \begin{subfigure}{0.32\linewidth}
		  \centering
		  \includegraphics[width=0.96\linewidth]{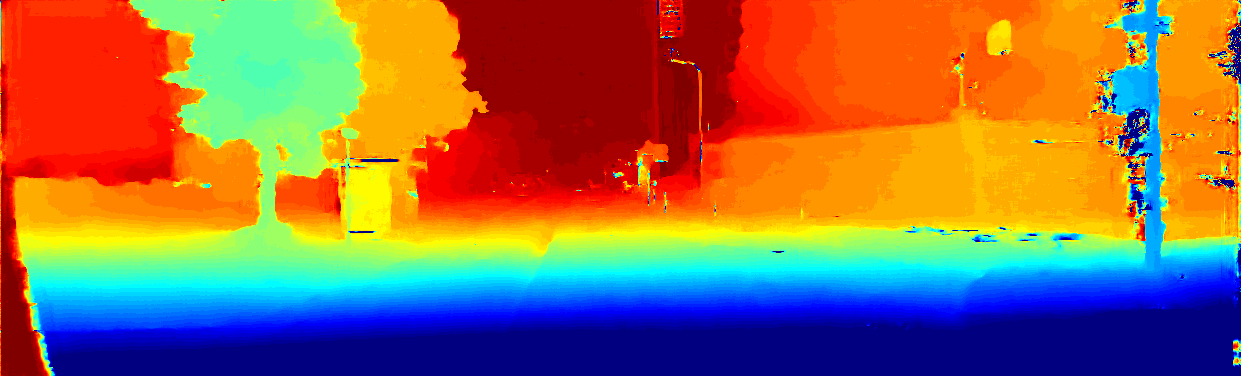}
	  \end{subfigure}
        \begin{subfigure}{0.32\linewidth}
		  \centering
		  \includegraphics[width=0.96\linewidth]{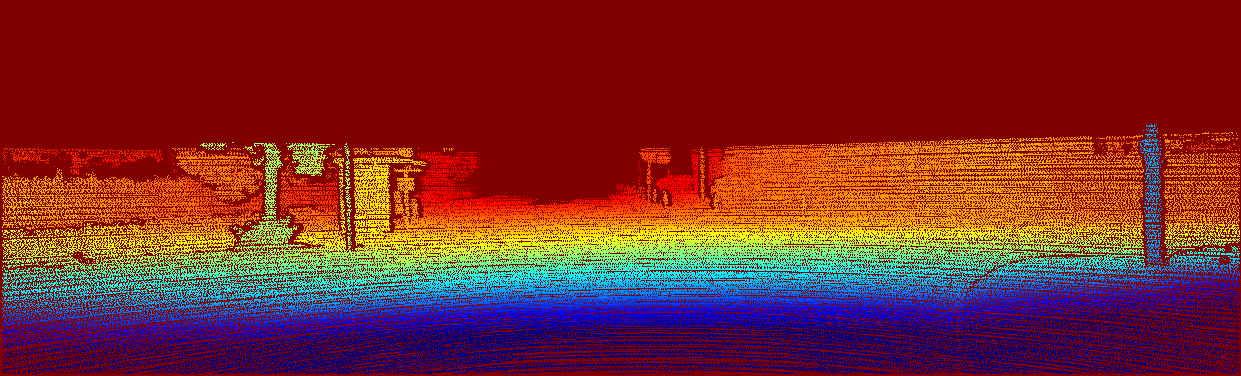}
	  \end{subfigure}
        \label{1-b}%文中引用该图片代号
	\end{subfigure}
    \centering
	\begin{subfigure}{0.96\linewidth}
		\begin{subfigure}{0.32\linewidth}
		  \centering
		  \includegraphics[width=0.96\linewidth]{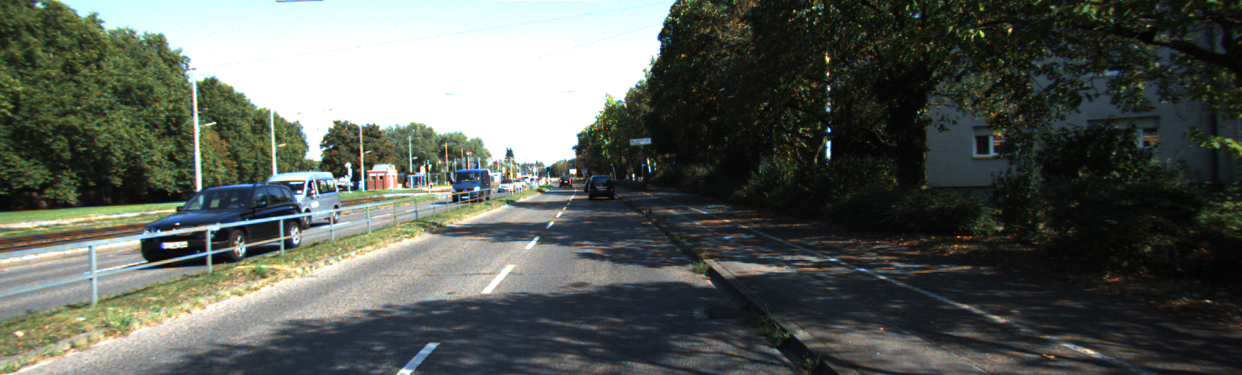}
	  \end{subfigure}
        \begin{subfigure}{0.32\linewidth}
		  \centering
		  \includegraphics[width=0.96\linewidth]{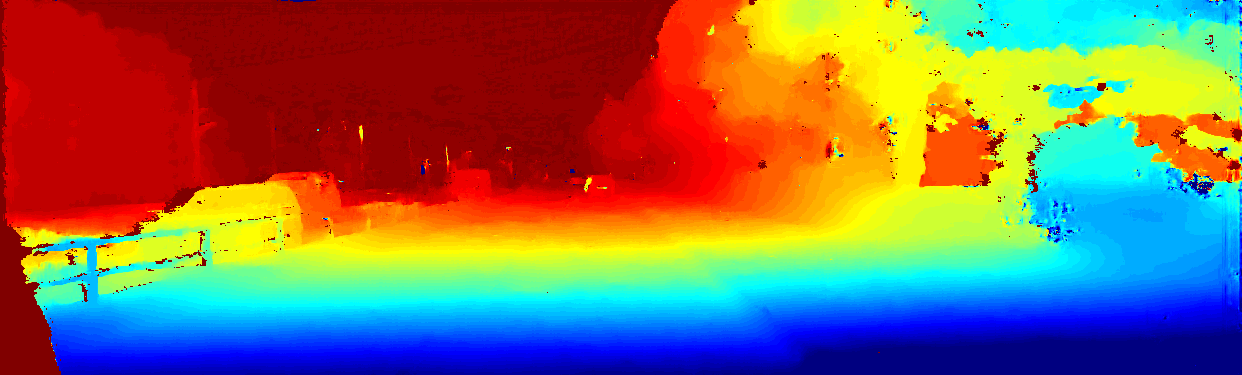}
	  \end{subfigure}
        \begin{subfigure}{0.32\linewidth}
		  \centering
		  \includegraphics[width=0.96\linewidth]{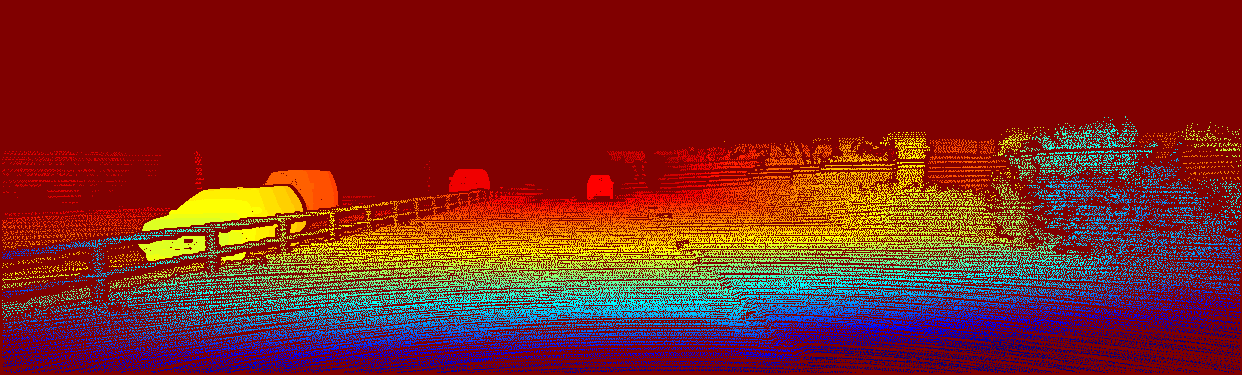}
	  \end{subfigure}
        \label{1-c}%文中引用该图片代号
	\end{subfigure}
	\centering
 
	\begin{subfigure}{0.96\linewidth}
		\begin{subfigure}{0.32\linewidth}
		  \centering
		  \includegraphics[width=0.96\linewidth]{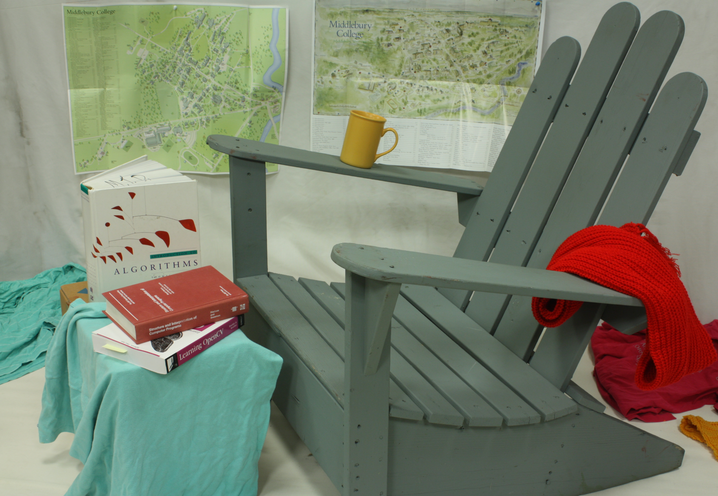}
            \caption{left}
	  \end{subfigure}
        \begin{subfigure}{0.32\linewidth}
		  \centering
		  \includegraphics[width=0.96\linewidth]{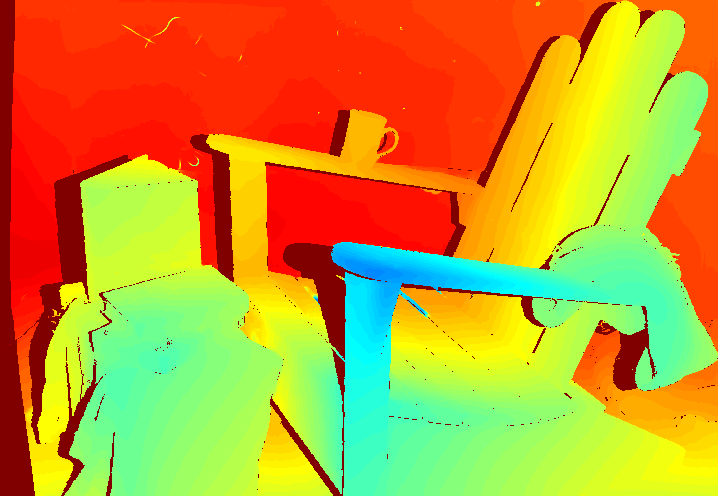}
            \caption{ours}
	  \end{subfigure}
        \begin{subfigure}{0.32\linewidth}
		  \centering
		  \includegraphics[width=0.96\linewidth]{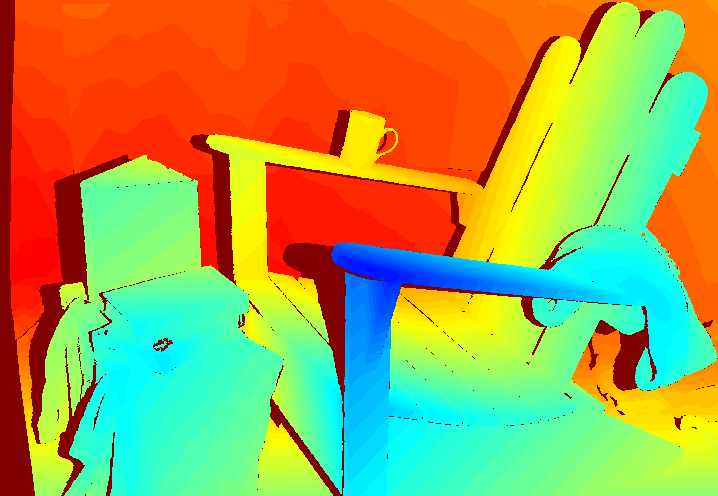}
            \caption{ground truth}
	  \end{subfigure}
        \label{1-d}%文中引用该图片代号
	\end{subfigure}
	\caption{Visualization of the results of cross generalization. From top to bottom, the samples of datasets ETH3D, KITTI 2012, KITTI 2015 and Middlebury 2014, respectively. The model is trained with the synthetic dataset Scene Flow without fine-tuning.}
	\label{Fig1}
\end{figure}
%%%%%%%%% BODY TEXT

\section{Introduction}
\label{sec:intro}
Stereo matching is a technique to identify the spatial corresponding pixels in binocular view and then the depth information can be recovered by triangulation, which is widely used in autonomous driving, 3D reconstruction and UAV navigation \cite{achtelik2009stereo}. Specifically, stereo matching is based primarily on a rectified stereo pair of images, in which the disparity map is generated by calculating the pixel distances between the corresponding pixels on the epipolar line. After utilizing the disparity map, depth information can be accurately calculated with camera parameters. Since the growth of deep learning and the advent of several pioneering works\cite{zbontar2016stereo, guo2019group, kendall2017end,chang2018pyramid}, learning-based methods have become the main tool for stereo matching.

CNN-based methods \cite{chang2018pyramid, zhang2019ga, xu2020aanet,lipson2021raft} generally utilize different feature backbone networks to extract features from left and right images, construct a matching cost volume with dual view features, and then aggregate the result of 4D cost volume with multiple 3D convolution. As compared with traditional stereo matching methods \cite{hirschmuller2005accurate,hamzah2010sum}, cost volume-based method have been achieved excellent performance on several scenarios. However, these methods generally suffer from limitations of disparity range because of the need to specify the maximum disparity when initializing the cost volume. And the area beyond the disparity range will be compressed to the maximum value, which is extremely incompatible for highway autonomous driving navigation\cite{badki2020bi3d}.

Transformer \cite{vaswani2017attention} has demonstrated superior performance than convolution on multiple computer vision tasks. Nevertheless, Transformer is not widely used in stereo matching, and one of the few representative work STTR \cite{li2021revisiting} is matched by the distribution of attention weights between left and right pixels with the uniqueness of the epipolar line constraint. It eliminates the limitation of fixed disparity range and introduces independent position information to increase feature description sensitivity in both viewpoints. However, when dot-product attention is performed on individual intra-row pixel sets, it does not characterize the local features of the image effectively. In addition, its multi-scale information is concentrated in the CNN backbone network and does not deliver a well translational equivariance in the transmit to Transformer part. These shortcomings may be the reason why STTR does not fine-tune well in real dataset. 

In this paper, we present a pure attention architecture to release the limitation of disparity while avoiding the singularity of local image features in a single-view for stereo matching. We attempt to expand the capability of attention such that it provides both denser and richer pixel-level features, while also using epipolar lines to offer strong matching constraints. Specifically, we assign different roles to self- and cross-attention, where self-attention is responsible for making the pixel features more comprehensive and cross-attention is responsible for recording the sensitivity of pixels for features and positions during the matching process. Thus, we combined with the recent window self attention \cite{liu2021swin} to develop a model for stereo matching, called AAUformer. The overall architecture of our model is shown in Fig \ref{Fig2}. For details, the main contributions of our model are as follows: (1) We propose AAUformer, which interprets stereo matching as separate feature extraction and correlation processes with only attention architecture. (2) We design a multi-scale U-shaped network structure with alternated attention blocks to achieve a coarse-to-fine matching process. (3) We introduce window self attention instead of intra-row self-attention to extract abundant semantic features at the pixel level before cross-matching. (4) We conduct experiments and demonstrate that our model can maintains comparable performance on both synthetic and real-world datasets.

%-------------------------------------------------------------------------
\section{Related Work}
\subsection{Stereo Matching }
Stereo matching has been a popular research topic for 3D vision due to its relevance to stereo depth estimation \cite{laga2020survey}. Traditional stereo matching methods basically consist of the following four steps \cite{hosni2012fast,liang2018learning,min2011revisit}: cost-computation, cost-aggregation, disparity-optimization, and post-processing steps. Although traditional methods have made significant progress, they still  have challenges in some non-regular, untextured regions.\par
MC-CNN  \cite{zbontar2016stereo} is the first to use a learning approach to solve the stereo matching problem, which builds a convolutional network to extract region block features and then aggregates the costs across regions. However, it uses many of the traditional algorithms for post-processing, which is not strictly an end-to-end network. In order to achieve end-to-end learning, GC-Net \cite{kendall2017end} constructs a four-dimensional cost volume to combine the left and right feature maps, and adopts soft argmin for the final regression disparity. Since then, the work based on the cost volume has become the main method for stereo matching and has achieved excellent performance on the public dataset. These improvement works can be divided into three main categories optimization of feature extraction, optimization of cost volume initialization, and optimization of cost aggregation. For example, PSMNet \cite{chang2018pyramid} adds a spatial feature pyramid \cite{he2015spatial} to enhance environmental information at different locations and scales before initializing cost volume. ACVNet \cite{xu2022attention} suppresses a large amount of redundant information by initializing the cost volume with the correlation of attention mechanisms, and achieves a high level of performance with fewer parameters. AANet \cite{xu2020aanet} designs adaptive same-scale and cross-scale aggregation module to perform feature aggregation for cost volume, which ensures excellent performance while greatly boosting speed. Although the cost volume approach performs very efficiently, it always suffers from a limitation of disparity, therefore, a few work started to research disparity-free for end-to-end stereo matching. Bi3D \cite{badki2020bi3d}, which converts the stereo matching into a series of binary classifications, and then performs disparity by an arbitrary coarse quantization method. Even though Bi3D is capable of defining the depth range flexibly, it requires constant fitting, and the wider the range, the longer time it will take. STTR \cite{li2021revisiting}revisits stereo matching from a sequence-sequence perspective, which uses attention to match the left and right pixels on the epipolar line. It translates two-dimensional image feature matching into one-dimensional sequence pairing to achieve true disparity-free matching, but it doesn't perform great in real scenes. Thus, our model uses the same matching mechanism, with the difference that we focus more on the extraction and correlation of features at multiple scales and use different attention to separate them for outstanding presentation. The final results show that our model achieves much better performance in a relatively short time.
\subsection{Vision Transformer}
\begin{figure*}[htbp]
	\begin{subfigure}{0.7\linewidth}
		\centering
		\includegraphics[width=0.9\linewidth]{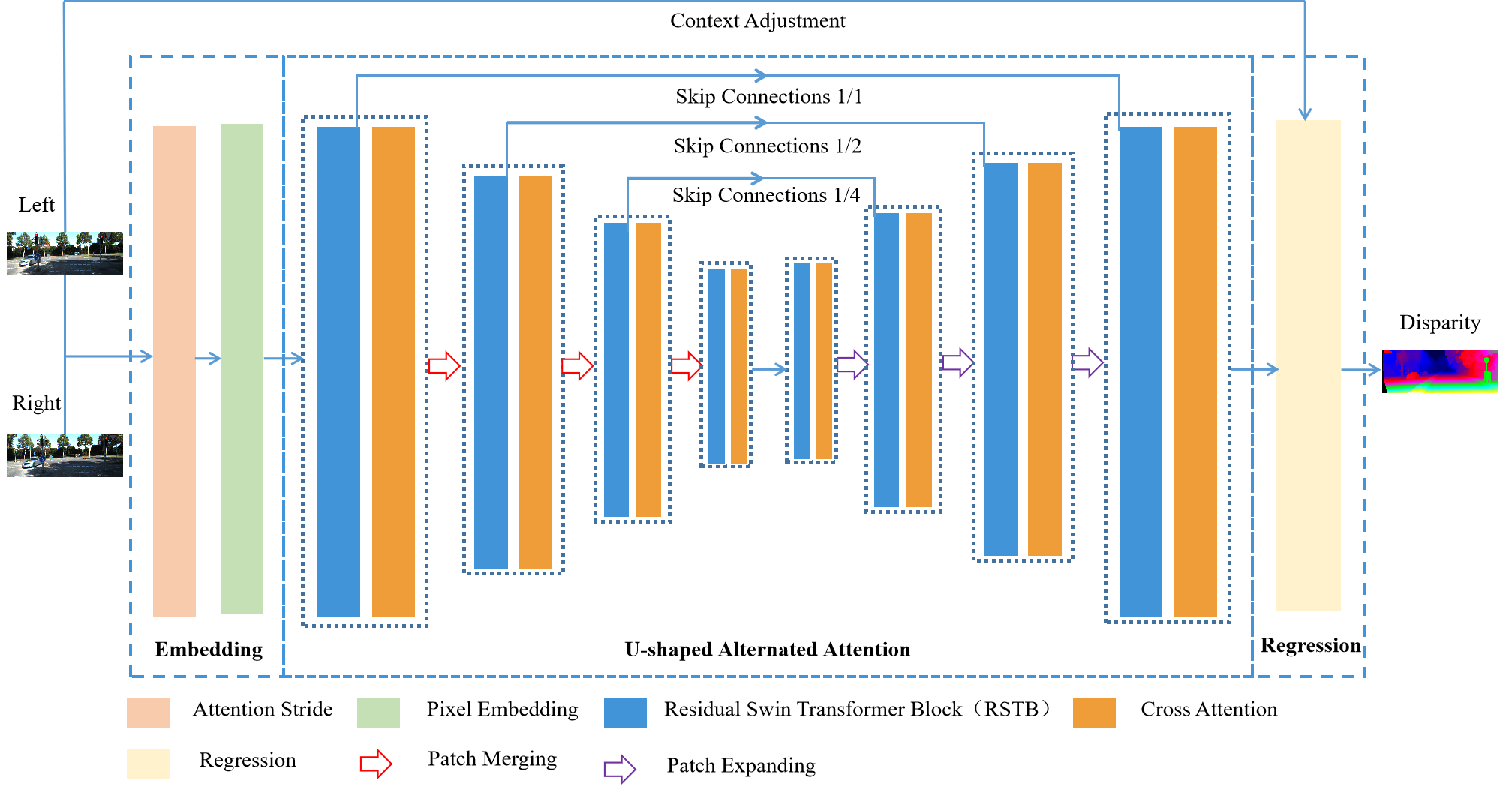}
		\caption{Network overview}
		\label{Fig2-1}%文中引用该图片代号
	\end{subfigure}
	%\qquad
	%让图片换行，
        \begin{subfigure}{0.4\linewidth}
    	\begin{subfigure}{0.9\linewidth}
    		\includegraphics[width=0.9\linewidth]{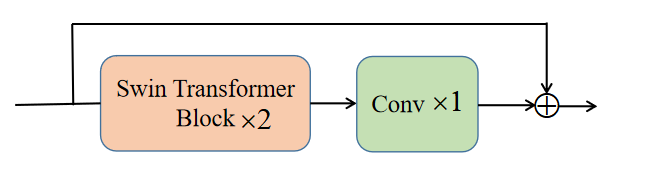}
    		\caption{Residual Swin Transformer Block (RSTB)}
    		\label{Fig2-2}%文中引用该图片代号
    	\end{subfigure}
    	\begin{subfigure}{0.9\linewidth}
    		\includegraphics[width=0.9\linewidth]{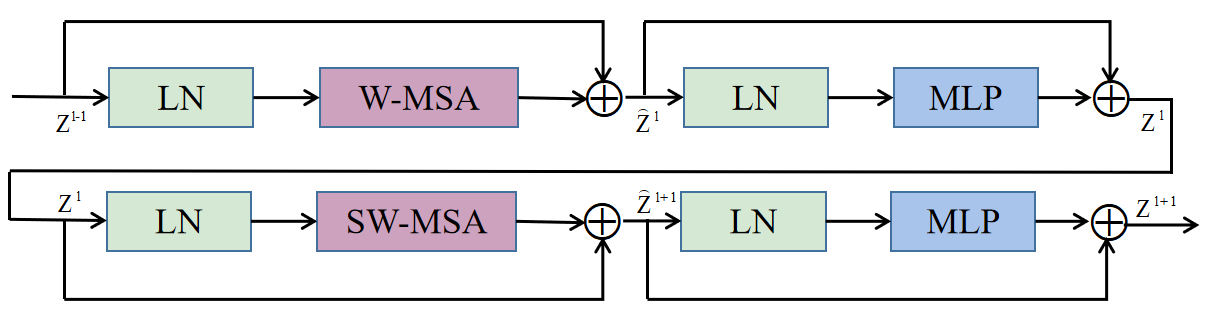}
    		\caption{Swin Transformer Block}
    		\label{Fig2-3}%文中引用该图片代号
    	\end{subfigure}
    	\begin{subfigure}{0.9\linewidth}
    		\includegraphics[width=0.9\linewidth]{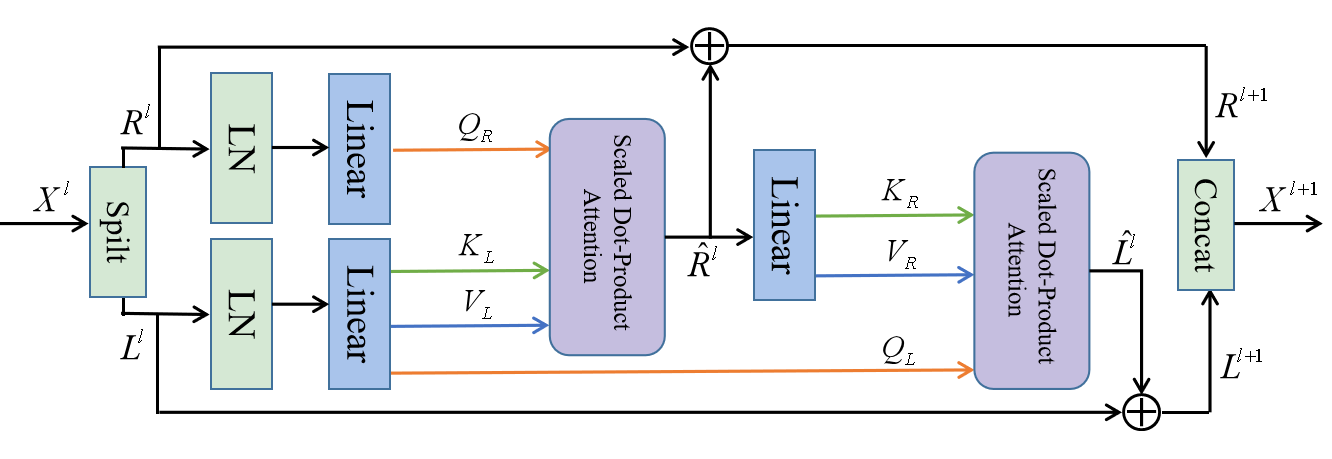}
    		\caption{Cross Attention}
    		\label{Fig2-4}%文中引用该图片代号
    	\end{subfigure}
         \end{subfigure}
	\caption{The overall network architecture and core components of AAUformer.}
	\label{Fig2}
\end{figure*}
Transformer \cite{vaswani2017attention} performs well in natural language processing tasks, as well as in computer vision tasks for its excellent self-attention mechanism. ViT \cite{dosovitskiy2020image} and DeiT \cite{touvron2021training} started the wave of using Transformer instead of convolutional networks to solve computer vision tasks. There are many applications of image classification \cite{dosovitskiy2020image, li2021localvit}, image segmentation \cite{wu2020visual, zheng2021rethinking, wang2021end}, object detection \cite{carion2020end, liu2020deep, dai2021up, touvron2021training}, and 3D vision \cite{lu2021geometry, wang2020position}. Recently, Swin Transformer \cite{liu2021swin} has appeared in various vision-related domains with its excellent capabilities. And much works have incorporated it in different approaches. Such as, by stacking multiple window attentions of Swin Transformer for feature extraction, Swin IR \cite{liang2021swinir} outperforms most other previous networks for super-resolution reconstructed images. Swin-UNet \cite{hatamizadeh2022swin} was constructed completely with Swin Transformer blocks for medical image segmentation, which also achieved the better performance for mainstream image segmentation datasets. The original authors of Swin proposed a more powerful model called Swin V2 \cite{liu2022swin}, which has a greater capacity and can be adapted to many high-resolution vision tasks.
%-------------------------------------------------------------------------
\section{Method}
\begin{figure*}[htbp]
	\centering
	\begin{subfigure}{0.18\linewidth}
		\centering
		\includegraphics[width=1\linewidth]{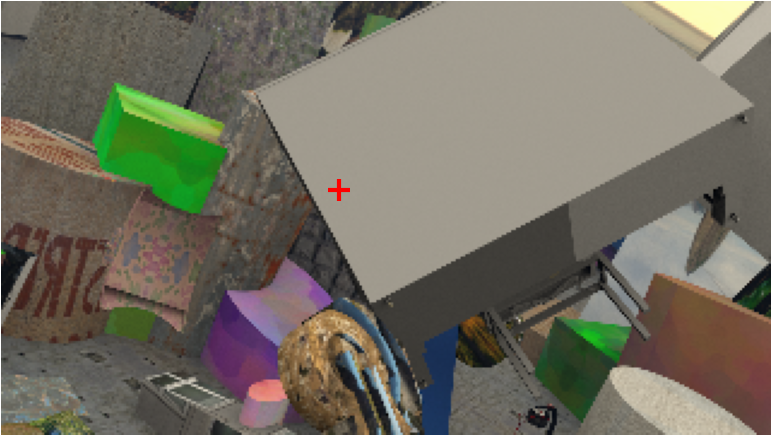}
		\caption{Left Image}
		\label{A-1}%文中引用该图片代号
	\end{subfigure}
	%\qquad
	%让图片换行，
	\begin{subfigure}{0.18\linewidth}
		\centering
		\includegraphics[width=1\linewidth]{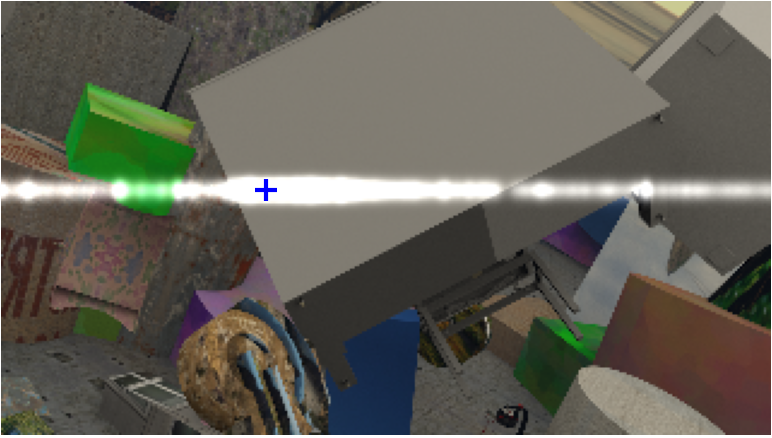}
		\caption{Cross 0 + Right Image}
		\label{A-2}%文中引用该图片代号
	\end{subfigure}
	\begin{subfigure}{0.18\linewidth}
		\centering
		\includegraphics[width=1\linewidth]{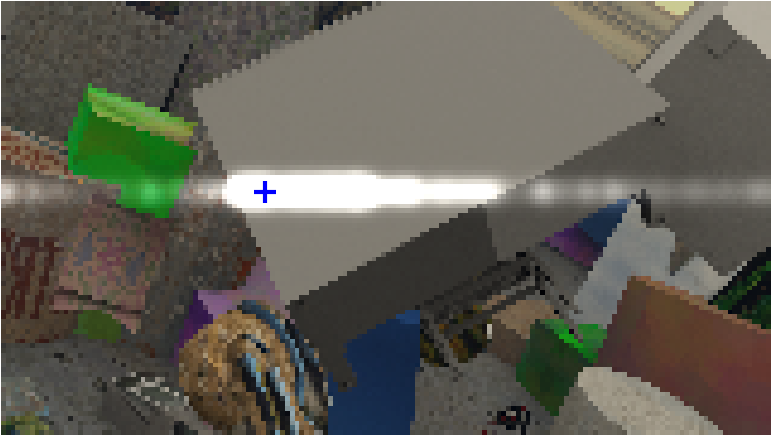}
		\caption{Cross 1 + Right Image}
		\label{A-3}%文中引用该图片代号
	\end{subfigure}
	\begin{subfigure}{0.18\linewidth}
		\centering
		\includegraphics[width=1\linewidth]{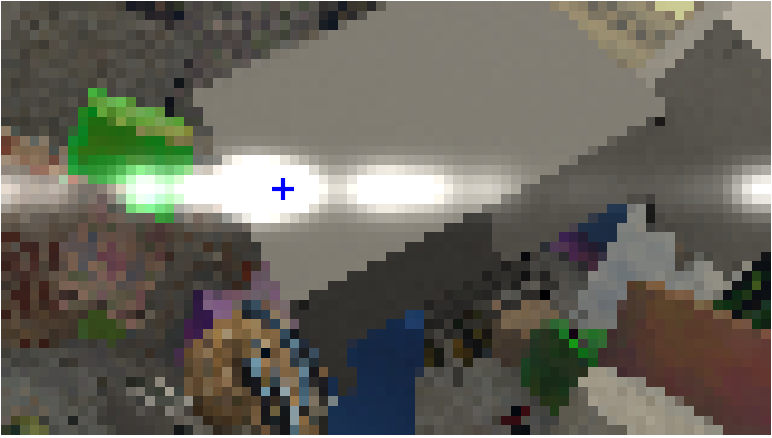}
		\caption{Cross 2 + Right Image}
		\label{A-4}%文中引用该图片代号
	\end{subfigure}
	\begin{subfigure}{0.18\linewidth}
		\centering
		\includegraphics[width=1\linewidth]{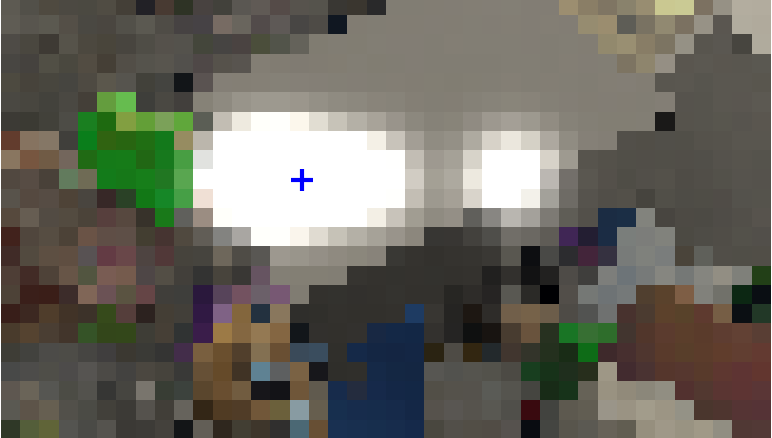}
		\caption{Cross 3 + Right Image}
		\label{A-5}%文中引用该图片代号
	\end{subfigure}
	\begin{subfigure}{0.18\linewidth}
		\centering
		\includegraphics[width=1\linewidth]{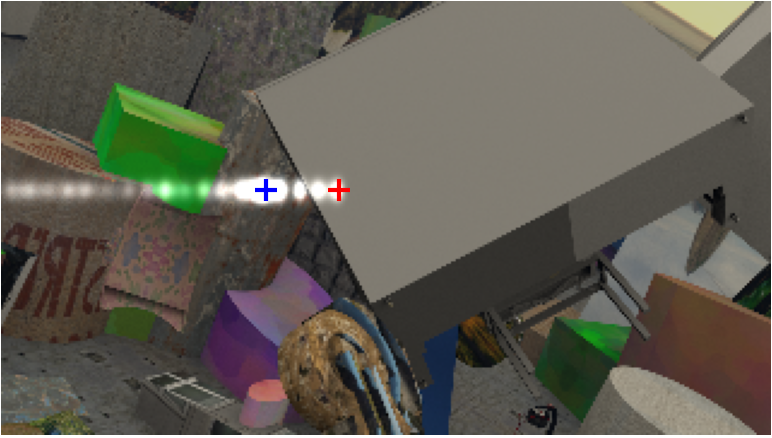}
		\caption{Cross 7 + Left Image}
		\label{A-10}%文中引用该图片代号
	\end{subfigure}
	\begin{subfigure}{0.18\linewidth}
		\centering
		\includegraphics[width=1\linewidth]{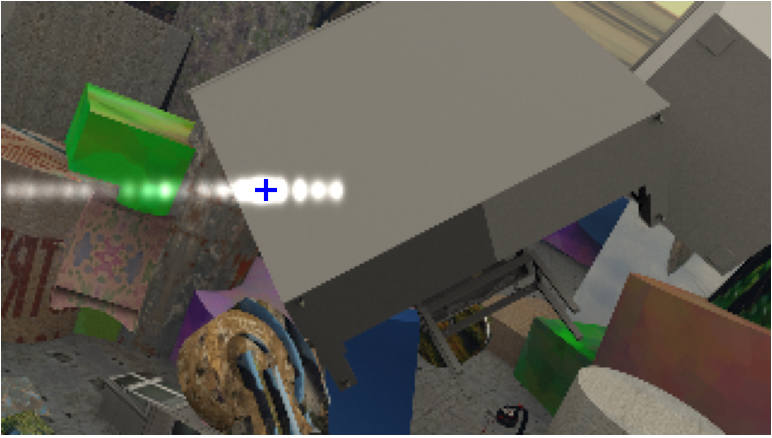}
		\caption{Cross 7 + Right Image}
		\label{A-9}%文中引用该图片代号
	\end{subfigure}
	\begin{subfigure}{0.18\linewidth}
		\centering
		\includegraphics[width=1\linewidth]{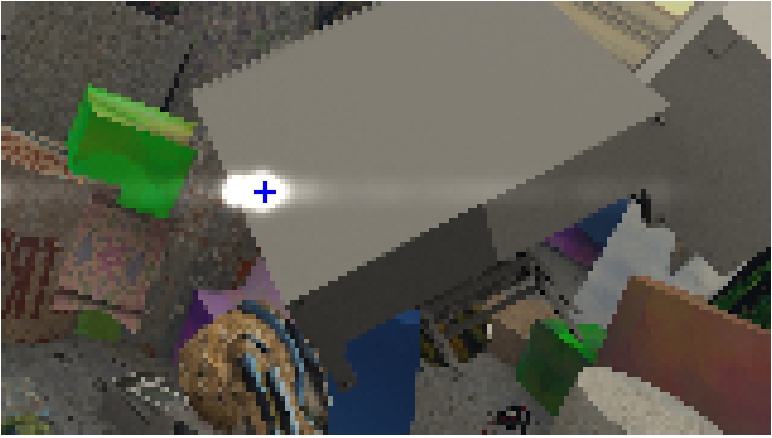}
		\caption{Cross 6 + Right Image}
		\label{A-8}%文中引用该图片代号
	\end{subfigure}
	\begin{subfigure}{0.18\linewidth}
		\centering
		\includegraphics[width=1\linewidth]{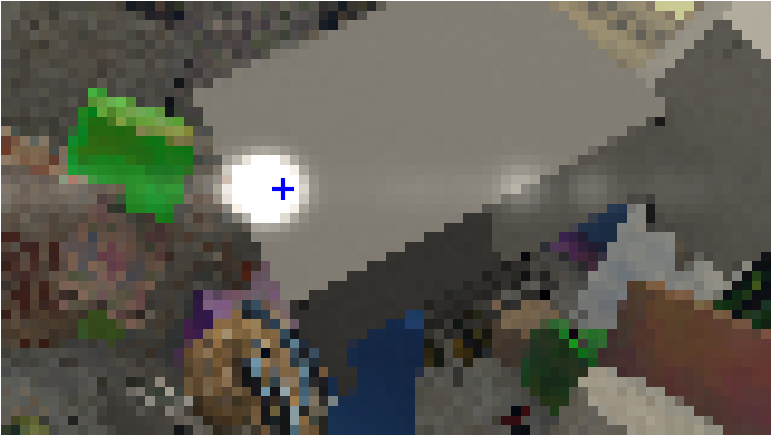}
		\caption{Cross 5 + Right Image}
		\label{A-7}%文中引用该图片代号
	\end{subfigure}
	\begin{subfigure}{0.18\linewidth}
		\centering
		\includegraphics[width=1\linewidth]{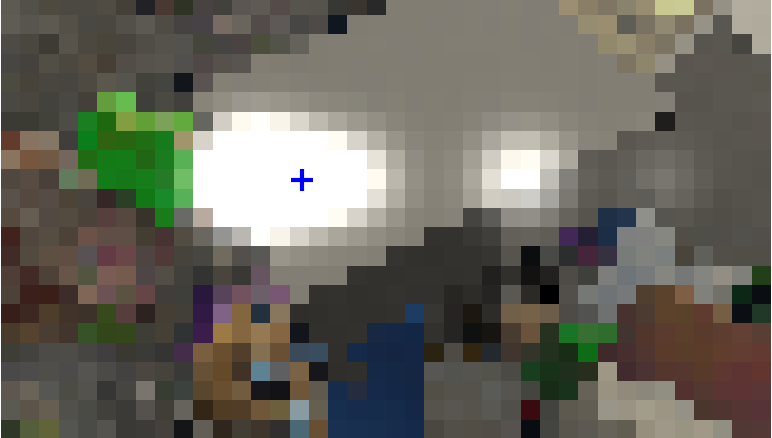}
		\caption{Cross 4 + Right Image}
		\label{A-6}%文中引用该图片代号
	\end{subfigure}
	\caption{Visual representation of cross-attention of U-shaped architecture. The red cross-hair on the left image represents the points of interest we selected. Highlighting represents the cross-attention weights of the epipolar lines located. The blue cross-hair represents the corresponding pixel matched by our network on the right image.}
	\label{appA}
\end{figure*}
\subsection{Network Architecture}
As shown in Fig \ref{Fig2-1}, the multi-scale U-shaped architecture is formed by alternated self-attention and cross-attention, where self-attention is acted by the RSTB. Attention will not change the size of the feature map, so the multi-level hierarchy is implemented with the patch merging layer and patch expanding layer. Besides, essential skip connections can contribute to feature fusion during up-sampling, and we perform ablation experiments in subsequent sections to study  its location and number for determine the best solution. For example, given the input left maps ${L\in(H,W,3)}$, the attention stride $s$, and the dimensions of the embedding $C$, the feature map at the i-th stage of the down-sampling process can be represented as ${L_i} \in \left( {\frac{H}{{s \times {2^i}}},\frac{W}{{s \times {2^i}}},{2^i} \times C} \right)$. In summary, our decision to introduce a U-shaped architecture is primarily based on two factors: first, the actual binocular images are not as perfect as the synthetic dataset, and often have aberrations, large viewpoint shifts, and some environmental noise, thus we want to extract invariant features \cite{lowe2004distinctive} by multi-scale to enhance the robustness of hard-to-discriminate regions. As shown in Fig \ref{appA}, we visualize the coarse-to-fine matching process for the corresponding pixels of the duplicated region. At the beginning, we can observe that the highlighting around the target pixel is difficult to separate in Fig \ref{A-2}. By performing scale down and up, the remaining pixel features with scale invariance can be matched more finely. As shown in Fig \ref{A-9}, the final highlights are concentrated on the target pixel and have a sharp distinction between the surrounding pixels. Second, the U-shaped architecture can reduce the size of features during the transmission process, which can physically reduce flops of attention and improve the speed of the network.

\subsection{Epipolar Line Constraint}
The epipolar line constraint is a geometric constraint on the same feature points in two or more camera coordinate systems, which describes the projection transfer process between the epipolar and image planes. Thus, the epipolar line constraint is not defined in a separate camera coordinate system or a single-view, and we distinguish this constraint between self- and cross-attention, corresponding to feature extraction and correlation. Specifically, for the point ${p}$ in the world coordinate system, assuming that the corresponding feature points in the left and right camera coordinate systems are ${p_r}$ and ${p_l}$, the polar line constraint can be described as follows
\begin{equation}
    p_l^T{K^{ - T}}t^\wedge R{K^{ - 1}}{p_r} = 0
\end{equation}
where, ${p_r,p_l}$ are homogeneous coordinate expressions. $K$ represents the intrinsics of the camera. $R, t$ are the rotation and translation matrix of the camera pose, respectively. $t^\wedge$ represents the antisymmetric matrix.
\subsection{Feature Extraction}
In traditional stereo matching algorithms, the neighborhood features around pixels are necessary. For example, SAD \cite{hamzah2010sum}, SGBM \cite{hirschmuller2005accurate} both begin with coarse matching by comparing region block features and then perform pixel-by-pixel matching. Inspired by this, we take the residual Swin Transformer block (RSTB) as our self-attention module to extract semantic features at the pixel level of the single-image and share weights in left and right images. As shown in Fig \ref{Fig2-2}, RSTB is a residual block consisting of Swin Transformer blocks and convolutional layer. The convolutional layers and residual connections in the module can strengthen the equivariance between the features of different blocks \cite{liang2021swinir}. The key components of RSTB are window self-attention (WSA) and shifted window self-attention (S-WSA).  Compared with intra-row self-attention (ISA) \cite{li2021revisiting}, given the image of $(h, w)$, the WSA partitions the image into $\frac{h}{M} \times \frac{w}{M}$ non-overlapping windows of $M \times M$, and performs self-attention within local windows, local WSA breaks the row limitations of the ISA and increases the amount of domain information. Additionally, we consider that WSA may focus more on local features and ignore the distant pixel on the epipolar lines, so we retain the shifted window mechanism to enhance the exchange of pixel information between windows. The computational complexity of ISA and WSA are 
\begin{equation}
    \Omega \left( {ISA} \right) = 4hw{C^2} + 2h{w^2}C
\end{equation}
\begin{equation}
    \Omega \left( {WSA} \right) = 4hw{C^2} + 2{M^2}hwC
\end{equation}
where, C is the number of channels in the feature map, M is the window size generally set to 7 \cite{liu2021swin}. The WSA is able to reduce the computational complexity compared to the ISA by scaling the $M$.

Furthermore, there is a significant component of Transformer that is the position embeddings, and the RSTB applies two types of position embeddings. One is the fully learnable absolute position embeddings \cite{gehring2017convolutional}, which is directly added to the embedding vector at the begining, and it contains translation invariance, monotonicity, and asymmetry \cite{wang2020position}, which can be used to learn more flexible features. Another one is learnable relative position encoding bias \cite{liu2022swin}, which is used in the attention matrix to record sensitive information at different levels of position.  With these two learnable position embeddings, the network will be able to learn more deep relevant features. In general, texture is a repeated local mode and arrangement rule in the image, and texture feature is a quantitative result of the local gray scale variation. 
\begin{figure}[htbp]
     \centering
     \includegraphics[width=1\linewidth]{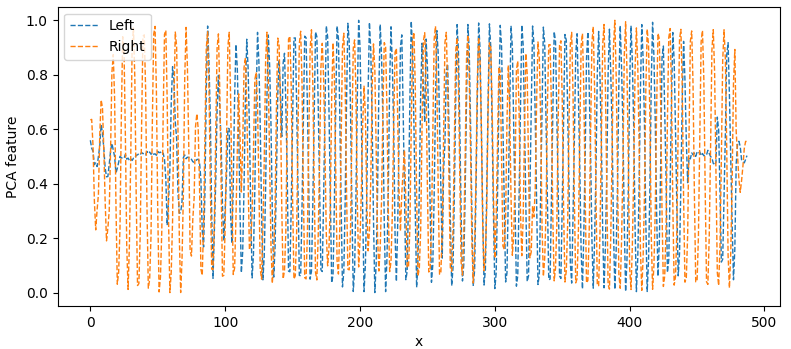}
     \caption{The PCA feature curves of the tensor from last self-attention. We select all pixels with row=30, and the dimensions of both left and right feature maps are (1,488,136,48), and the inference data samples are from Scene Flow. More PCA results can be referenced in Appendix \ref{pca_result}.}
     \label{PCA}%文中引用该图片代号
\end{figure}

In order to clearly understand this abundant local texture feature, we use principal component analysis(PCA)\cite{shlens2014tutorial} to characterize the feature map of the last layer of self-attention. As shown in Fig \ref{PCA}, we observe that almost all the features vary approximately periodically, and this variation trend corresponds to the texture features that we extracted. 
\subsection{Feature Correlation}
The features provided by the RSTB are derived from the inherent information in the single view.  For the stereo image, the left and right view features have strong similarity, which is beneficial for constructing stable contact for matching. To correlate the features on both views, twice attentions are performed in the cross-attention module with epipolar constraints. The specific operations are shown in Fig \ref{Fig2-4} and the details calculation process is as follows
\begin{equation}
    {R^l},{L^l} = Spilt\left( {{X^l}} \right)
\end{equation}
\begin{equation}
    {Q_R} = Linear\left( {LN\left( {{R^l}} \right)} \right)
\end{equation}
\begin{equation}
   {Q_L},{K_L},{V_L} = Linear\left( {LN\left( {{L^l}} \right)} \right)
\end{equation}
\begin{equation}
    {\hat R^l} = MSA\left( {{Q_R},{K_L},{V_L}} \right)
\end{equation}
\begin{equation}
      {K_R},{V_R} = Linear\left( {{{\hat R}^l}} \right)
\end{equation}
\begin{equation}
    {\hat L^l} = MSA\left( {{Q_L},{K_R},{V_R}} \right)
\end{equation}
\begin{equation}
    {X^{l + 1}} = Concat\left( {{R^l} + {{\hat R}^l},{L^l} + {{\hat L}^l}} \right)
\end{equation}
where ${X^l}$, ${X^{l + 1}}$ is the output of the ${l,l+1}$ cross-attention layer, respectively, ${R^l},{L^l},{\hat R^l},{\hat L^l}$ is an intermediate variable generated during the computation, $LN()$ denote layer normalization, ${Linear()}$ represents linear projection, and $MSA(Q,K,V)$ is the multi-headed dot-product attention. 

In cross-attention, position embeddings is also essential, and we also retained the independent relative position coding \cite{li2021revisiting, dai2019transformer}, where each pixel on the epipolar line contains a data vector ${e_d}$ and a position vector ${e_p}$ based on the pixel's position, calculating attention separately for data-to-data attention, data-to-position attention, and position-to-data attention, and then summed. In this case, attention between the $i$-th and $j$-th can be computed as
\begin{equation}
    {a_{i,j}} = [{e_{d,i}} \otimes {e_{d,j}}] + [{e_{d,i}} \otimes {e_{p,j-i}}] + [{e_{p,i - j}} \otimes {e_{d,j}}]
\end{equation}
where$[. \otimes .]$ represents two vectors for dot product attention. $ {e_{p,i - j}}$ represents the relative position between $i,j$. It is needed to distinguish that $ {e_{p,i - j}}$ and $ {e_{p,j - i}}$ are not equal. 

\begin{figure}[htbp]
     \centering
     \includegraphics[width=1\linewidth]{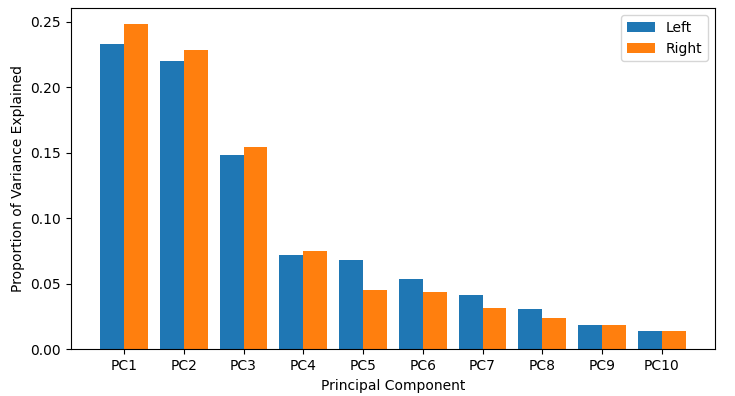}
     \caption{The component distribution percentages of the final feature map after PCA compression. The inference data samples are from Scene Flow.}
     \label{PCA_C}%文中引用该图片代号
\end{figure}

 To verify the correlation between the left and right features, we use PCA to conduct a comparison of the same features in the left and right views, and the experimental results are shown in Fig \ref{PCA_C}. It is obvious that the feature components in the left and right feature maps are pretty similar.
\subsection{Disparity and Occlusion Regression}
In this section, we refer to the regression structure of STTR \cite{li2021revisiting}, which consists of three sections: Optimal Transport \cite{cuturi2013sinkhorn}, Raw Regression, and Context Adjustment Layer \cite{yu2018wide}. First, the final attention matching matrix is iterated using Optimal Transport with soft assignment and differentiability, in order to increase the cost of some pixels that cannot be matched due to occlusion. After that, Raw Disparity and Occlusion Regression are performed to determine the matching pixel with the maximal attention value for each pixel, and a window of linear interpolation is performed to refine the sub-pixel level corresponding pixels. And then, attention stride is used throughout the training process due to hardware limitations. Naturally, a super-resolution reconstruction of the low-resolution Raw Disparity and Occlusion are performed using the Context Adjustment Layer, and the final image features are connected through multiple residual connection blocks in the Context Adjustment Layer can unite the initial left image information to complete the whole regression process. Details of the Loss function for the regression part are in Appendix \textcolor{red}{A}.

\section{Experiments and Analysis}
\textbf {Datasets and Evaluation Metrics:} we conducted our experiments on four popular stereo datasets, including Scene Flow \cite{mayer2016large}, KITTI \cite{menze2015joint,geiger2012we}, ETH3D \cite{schops2017multi}, and Middlebury2014 \cite{scharstein2002taxonomy}. Scene Flow is a synthetic datasets with random objects, and we select the FlyingThings3D subset, which provides 21,818 image pairs for training with a resolution of 960 × 540, as well as provides dense disparity maps and masks of occluded regions. KITTI dataset consists of street scenes and their depth information obtained by LIDAR. It provides a relatively sparse disparity map with the 2012 and 2015 subsets. Middlebury2014 is an indoor scene dataset, providing dense disparity and mask information with occlusion. We chose half and quarter size images for testing, only 15 image pairs in total. We use 3 px Error (percentage of errors larger than 3 px) and EPE (absolute error) to evaluate the non-occluded region. It should be noted that the pairs of images provided in these datasets above are rectified to approximately satisfy the epipolar line constraint. Besides, we supplement the generalization comparison tests for different alignment levels in Appendix \ref{Alignment}.

\textbf {Implementation Details:} we implement our approach in Pytorch and using AdamW as the optimizer with decay of 1e-4. The initial learning rate of Transformer is set to le-4, and the learning rate of the context adjustment layer in regression is 2e-4. Besides, the window size is set to 7 for Swin Transformer blocks. The device we used for training was an NVIDIA RTX3090 GPU with 24GB memory. Since the Scene Flow provides a relatively large number of images, we choose it as a fundamental training dataset for our model. We set up the training for 100 epochs, each epoch is about 3.5 hours and we use attention stride of s=3 during training.  
\subsection{Ablation Studies}
In this section we perform a series of ablation experiments on the Scene Flow dataset to evaluate the performance of our model with different settings. Specifically, including the number of WSA and S-WSA blocks, skip connections, and position embeddings.

\textbf{Effect of the number of WSA and S-WSA blocks:} the number of  WSA and S-WSA blocks in each self-attention is set to 2, 4 and 6 for base-model, middle-model and large-model. Each model is restarted to train 100 epochs and then validated on the test dataset. As listed in Table \ref{tab1}, The increase in number can reduce the 3px Error of 0.11 and the EPE of about 0.02, which can demonstrate the potential of large models to improve the model performance. At the same time, it also brings an increase in the number of parameters and computational complexity, So it is necessary take the model scales into consideration in specific scenarios.
\begin{table}
  \centering
  \begin{tabular}{c c c c c}
    \toprule
    Model & 3 px Error& EPE&Params[M]&Flops[G] \\
    \midrule
    Base &	   0.87	 & 0.35 &	10.8M &	148.5G \\
    Middle &	0.79 &	0.34 &	20.3M &	206.1G \\
    Large &	\textbf{0.78} &	\textbf{0.33} &	29.8M &	263.6G \\
    \bottomrule
  \end{tabular}
  \caption{Ablation study on the impact of the number of WSA and S-WSA blocks. where Base corresponds to the number of blocks consisting of WSA and S-WSA in each self-attention is 2, Middle corresponds to 4 and Large corresponds to 6. \textbf{Bold} is best.}
  \label{tab1}
\end{table}

\textbf{Effect of skip connections:} the skip connections in our model are added to 1/2, 1/4, and 1/8 resolution scales. We will conduct ablation experiments in terms of both the number and location of skip connections. Separately, we change the number of skip connections to 0, 1, 2, 3 to explore the effect of the number and also we choose skip connections to be added before or after the cross-attention module to explore the effect of the location . In order to make full use of the training resources, we first train the base model with 60 epochs and 0 number of skip connections, and then fine-tune the model with increasing the number of connections . As listed in Table \ref{tab2}, both the increase in the number of skip connections and the add of skip connections before cross-attention can improve the performance of the model. As a result, our model uses a number of skip connections of 3 and is added before the cross-attention module.

\textbf{Effect of position embeddings:} the position embeddings include absolute position embeddings and relative position embeddings, which are applied in our RSTB and cross-attention modules, respectively. We designed four schemes to explore the impact of position embeddings on our network, namely, using neither position embeddings, using only absolute position embeddings, using only relative position embeddings and using both position embeddings. As listed in Table \ref{tab3}, we find that while using the two position embeddings separately also enhances the performance of our model, the boost is better when used together. Therefore, we use both position embeddings in our formal model.

\begin{table}
  \centering
  \begin{tabular}{c c c c c}
    \toprule
    Skip  & \multicolumn{2}{c}{Before} & \multicolumn{2}{c}{After} \\
    \cline{2-3}
    \cline{4-5}
    Connections & 3 px Error & EPE & 3 px Error & EPE \\
    
    \midrule
    0 &	    1.25 &            0.46 &	             1.28 &	   0.47 \\
    1 & 	1.07 &	          0.41 &	              1.15 &	   0.44 \\
    2 &	    0.95 &	          0.37 &	              1.03 &	   0.40 \\
    3 &	    \textbf{0.87} &	          \textbf{0.35} &	              0.94 &	   0.37 \\
    \bottomrule
  \end{tabular}
  \caption{Ablation study on the impact of skip connections. Before: the add of skip connections before cross-attention. After: the add of skip connections after cross-attention. \textbf{Bold} is best.}
  \label{tab2}
\end{table}

\begin{table}
  \centering
  \begin{tabular}{c c c}
    \toprule
    Position embeddings & 3 px Error& EPE \\
    \midrule
    None &	   1.38	 & 0.51  \\
    Ape &	1.12 &	0.44  \\
    Rpe &   1.08 & 0.42 \\
    Ape+Rpe &	\textbf{0.87} &	\textbf{0.35}  \\
    \bottomrule
  \end{tabular}
  \caption{Ablation study on the impact of position embeddings. None: using neither position embeddings. Ape: using only absolute position embeddings. Rpe: using only relative position embeddings. Ape+Rpe: using both position. \textbf{Bold} is best.}
  \label{tab3}
\end{table}
\subsection{Cross Generalization Experiment}
\begin{figure}[htbp]
    \centering
    \begin{subfigure}{0.45\linewidth}
        \centering
        \includegraphics[width=0.8\textwidth]{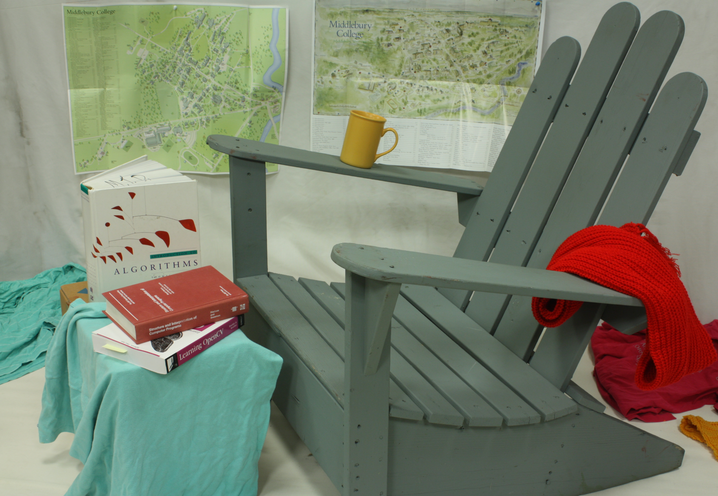}
        \caption{Middlebury}
    \end{subfigure}
    \centering
    \begin{subfigure}{0.45\linewidth}
        \centering
        \includegraphics[width=0.8\textwidth]{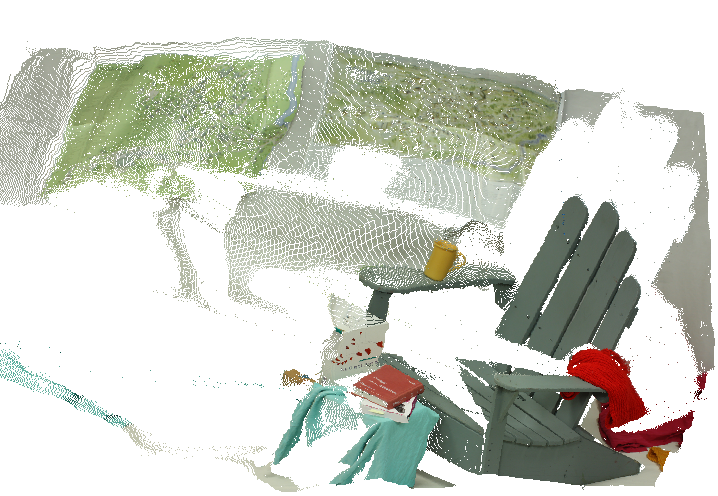}
        \caption{Middlebury-point cloud}
    \end{subfigure}
    \centering
    \begin{subfigure}{0.45\linewidth}
        \centering
        \includegraphics[width=0.8\linewidth]{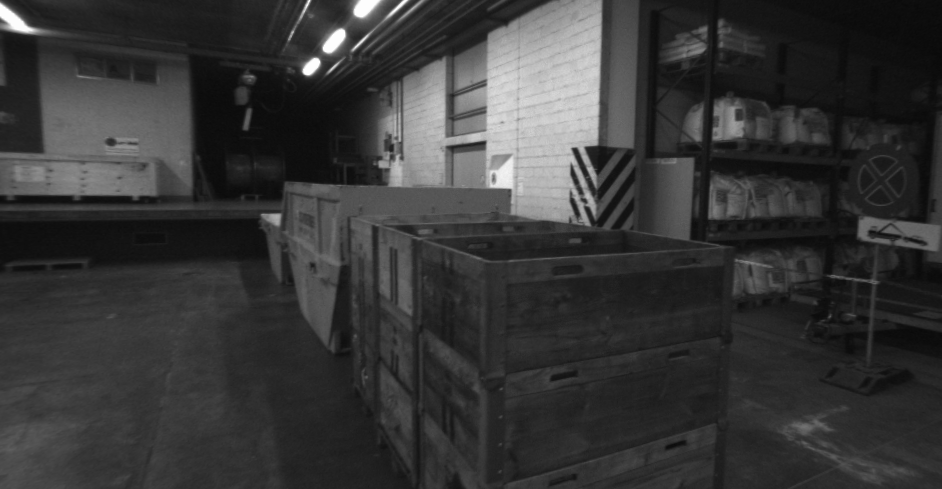}
        \caption{ETH3D}
    \end{subfigure}
    \centering
    \begin{subfigure}{0.45\linewidth}
        \centering
        \includegraphics[width=1\linewidth]{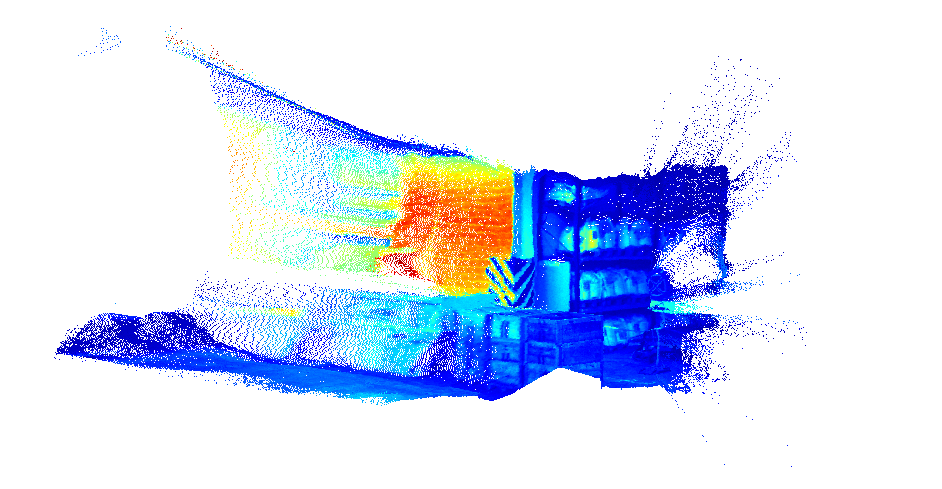}
        \caption{ETH3D-point cloud}
    \end{subfigure}
    \caption{Visualization of part of the point cloud for cross generalization results.}
    \label{Cross4.2}
\end{figure}

In this section we will conduct experiments to verify the cross generalization ability of our model, and we choose the model trained on Scene Flow to evaluate it directly on the other real-world dataset without  fine-tuning. The comparison results with the previous work evaluated are shown in Table \ref{tab4}. It can be seen that our model basically outperforms other comparison methods and we additionally present the visualization of the samples results as shown in Fig \ref{Fig1}. For the samples from the ETH3D and Middlebury datasets, we recovered the disparity map to a three-dimensional point cloud in space, as shown in Fig \ref{Cross4.2}. 
\begin{table}
  \centering
  \begin{tabular}{c c c c c c}
    \toprule
    \multirow{2}{*}{Method} & \multicolumn{2}{c}{Middlebury} & \multicolumn{2}{c}{KITTI} & \multirow{2}{*}{ETH3D}\\
    ~ & half & quarter &  2012 & 2015 & ~ \\
    \midrule
    CostFilter \cite{hosni2012fast} & 40.5 & 17.6 & 21.7 & 18.9 & 31.1 \\
    PatchMatch \cite{bleyer2011patchmatch} & 38.6 & 16.1 & 20.1 &17.2 & 24.1  \\
    SGM \cite{hirschmuller2007stereo} & 25.2 & 10.7 & 7.1  & 7.6 & 12.9 \\ 
    AANet \cite{xu2020aanet} & 14.5 & 12.8 & 10.1 & 12.4 & 9.8 \\
    GwcNet-g \cite{guo2019group} & 11.1 & 8.5 &  8.6 & 9.1 &8.9\\
    STTR \cite{li2021revisiting} & 15.5 & 9.7 & 8.7 & 6.7 & 17.2 \\
    PSMNet \cite{chang2018pyramid} & 15.8 & 9.8 & 6.0  &6.3 & 10.2\\
    FC-PSMNet \cite{zhang2022revisiting} &15.1 & 9.3 & 5.3 & 5.8 & 9.5\\
    CFNet \cite{shen2021cfnet} & 15.3 & 9.8 & 4.7 & 5.8 & 5.8\\
    GANet \cite{zhang2019ga} & 13.5 & 8.5  &5.5& 6.0&6.5   \\
    FC-GANet\cite{zhang2022revisiting} & 10.2 & 7.8 & 4.6 & 5.3 & 5.8\\
    Ours & \textbf{9.5} & \textbf{5.9} & \textbf{4.5} & \textbf{5.1} & \textbf{3.2}\\
    \bottomrule
  \end{tabular}
  \caption{Evaluation results of the generalization ability experiment. Inference results on KITTI 2012, KITTI 2015, Middlebury 2014 and ETH3D datasets with models trained on Scene Flow datasets without fine-tuning. Threshold error rate ($\%$) is adopted. \textbf{Bold} is best.}
  \label{tab4}
\end{table}

\subsection{Scene Flow Benchmark Result}
\begin{figure}[htbp]
    \centering
    \begin{subfigure}{0.32\linewidth}
        \centering
        \includegraphics[width=0.95\textwidth]{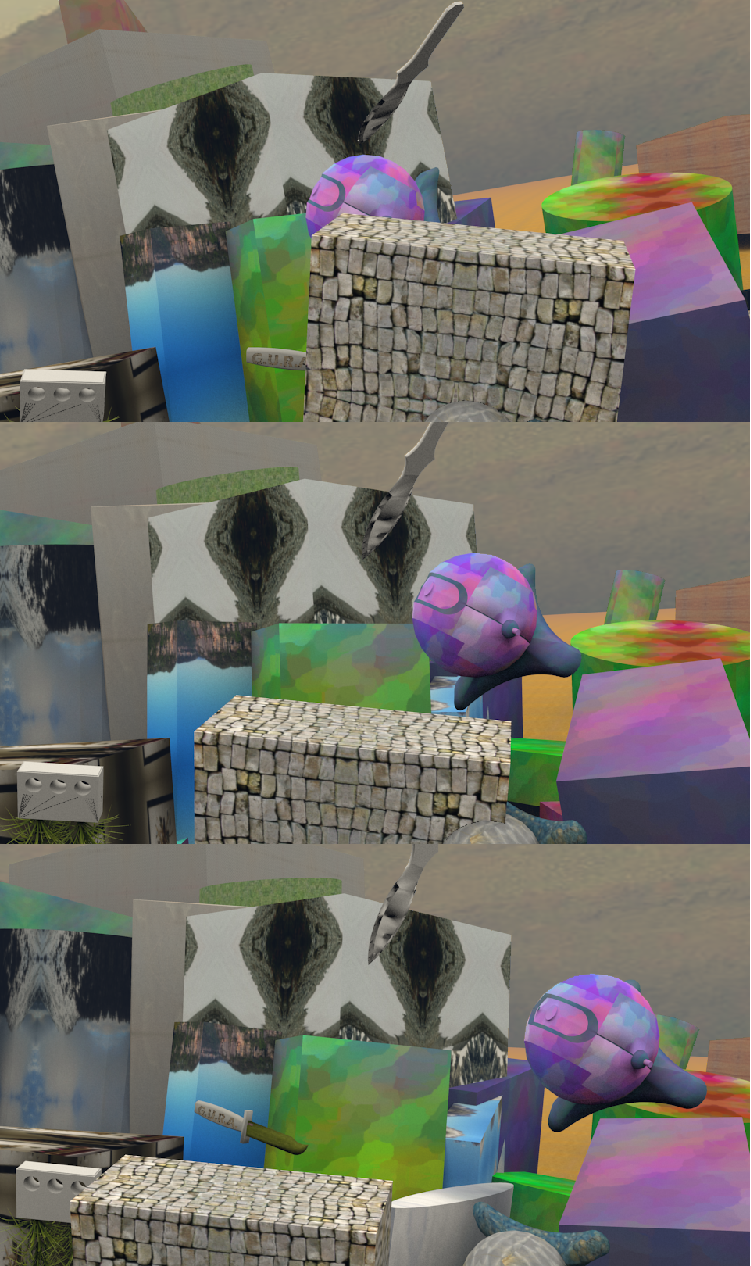}
        \caption{left}
        \label{5-1}
    \end{subfigure}
    \centering
    \begin{subfigure}{0.32\linewidth}
        \centering
        \includegraphics[width=0.95\linewidth]{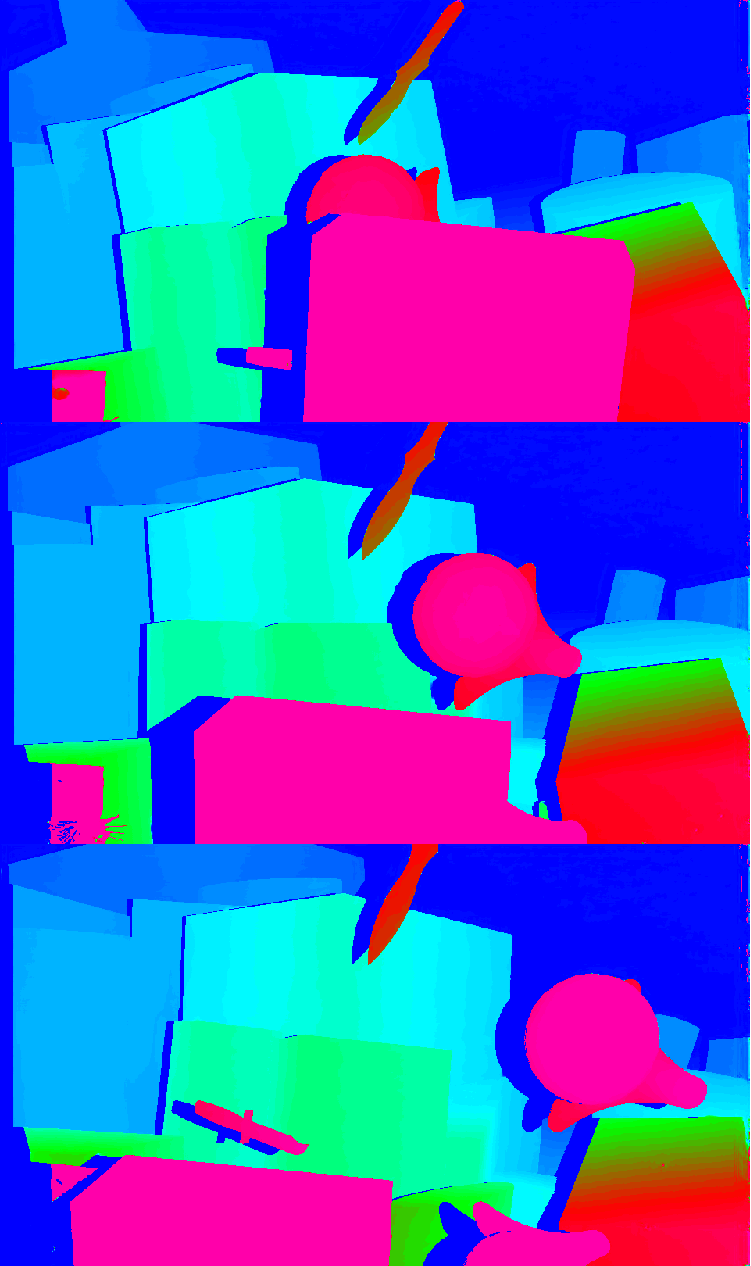}
        \caption{ours}
        \label{5-2}
    \end{subfigure}
    \centering
    \begin{subfigure}{0.32\linewidth}
        \centering
        \includegraphics[width=0.95\linewidth]{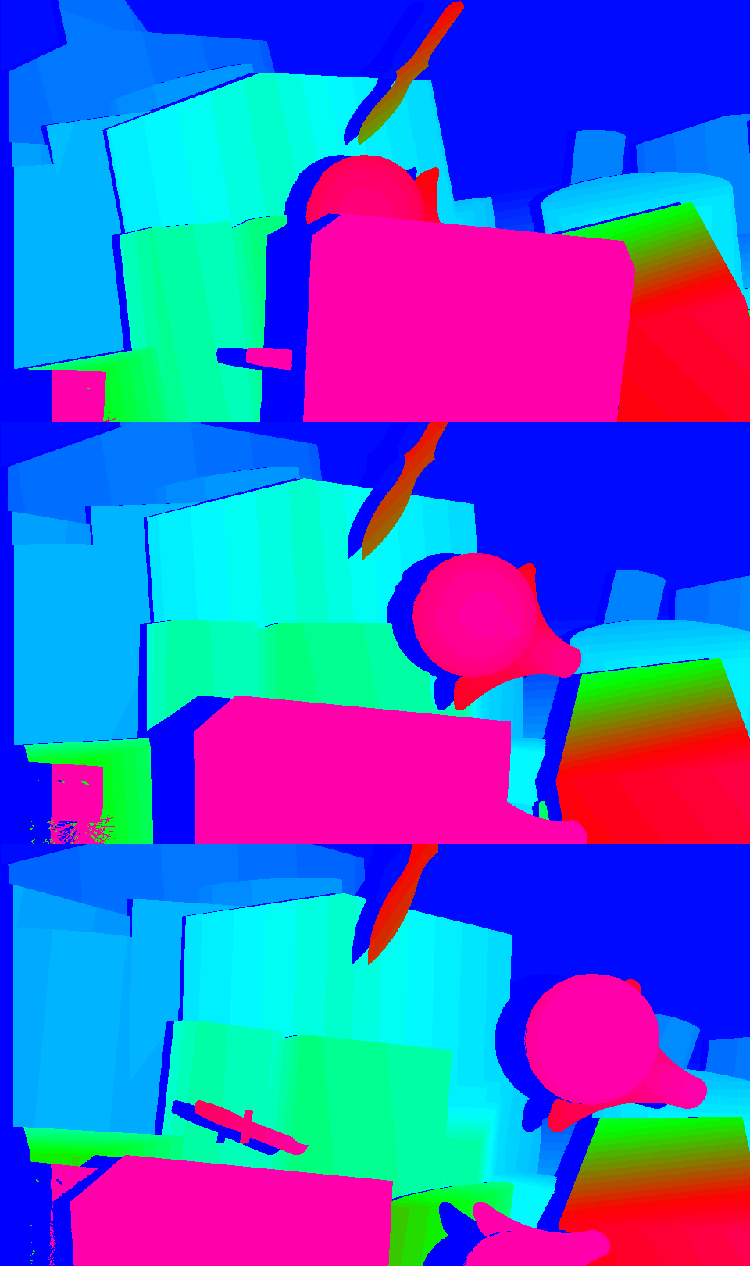}
        \caption{ground truth}
        \label{5-3}
    \end{subfigure}
    \caption{The visualization of inference results of three continuous frames of samples from Scene Flow.}
    \label{SceneFLow_all}
\end{figure}

As shown in Table \ref{tab5}, we set the pixel metrics for two regions, the regions with disparity less than 192 and all non-occluded regions, considering that most of the previous work required setting a fixed maximum disparity value. The results show that in both case, our model is still able to outperform the previous work and achieves state-of-the-art. As shown in Fig \ref{SceneFLow_all}, we visualize the inference results for three continuous frames of samples, and it can be seen that our predictions are pretty close to the ground truth.  In addition, we performed some comparative experiments on Scene Flow for our model computational complexity and inference time can be shown in Appendix \ref{Complexity}.

\begin{table}
  \centering
  \begin{tabular}{c c c c c}
    \toprule
    \multirow{2}{*}{Model} & \multicolumn{2}{c}{disp$<$192} & \multicolumn{2}{c}{All pixels} \\
    \cline{2-3}
    \cline{4-5}
    ~ & 3 px Error & EPE & 3 px Error & EPE \\
    
    \midrule
    PSMNet\cite{chang2018pyramid} &	    2.87 &            0.95 &	              3.31 &	   1.25 \\
    GwcNet-g\cite{guo2019group} &  	1.57 &	          0.48 &	              2.09 &	   0.89 \\
    HITNet\cite{tankovich2021hitnet} &	    2.21 &	          0.36 &	              2.69 &	   0.75 \\
    GANet-11\cite{zhang2019ga} &	    1.60 &	          0.48 &	              2.19 &	   0.97 \\
    AANet\cite{xu2020aanet} &	        1.86 &	          0.49 &	              2.38 &	   1.96 \\
    Bi3D\cite{badki2020bi3d} &	        1.70 &	          0.54 &	              2.21 &	   1.16 \\
    LEAStereo\cite{cheng2020hierarchical} &	    2.60 &	          0.78 &	              3.17 &	   1.12 \\
    
    STTR\cite{li2021revisiting} &	        1.13 &	          0.42 &	              1.26 &	   0.45 \\
    Ours &	        \textbf{0.74} &	          \textbf{0.32} &	              \textbf{0.87} &	   \textbf{0.35} \\
    \bottomrule
  \end{tabular}
  \caption{Comparisons with state-of-the-art methods on Scene Flow dataset, All pixels represents the non-occluded region, (disp$<$192) is the region of All pixels in which the disparity value is less than 192. \textbf{Bold} is best.}
  \label{tab5}
\end{table}
\begin{table}
  \centering
  \begin{tabular}{c c c c}
    \toprule
    \multirow{2}{*}{Model}&  \multicolumn{3}{c}{Noc($\%$)} \\
    ~ & D1-bg & D1-fg & D1-all \\
    \midrule
    AANet\cite{xu2020aanet} &	                    1.80 &	              4.93 &	   2.32 \\
    ACVNet\cite{xu2022attention} &  	             \textbf{1.26} &	              2.84 &	   1.52\\
    GwcNet-g\cite{guo2019group} &	              1.61 &	             3.49 &	   1.92 \\
    Bi3D\cite{badki2020bi3d} &	                 1.79 &	              3.11 &	  2.01 \\
    STTR\cite{li2021revisiting} &	        	          1.70 &	              3.61 &	   2.01 \\
    LaC+GANet\cite{liu2022local} &  \textbf{1.26} & 2.64 & \textbf{1.49}\\
    LEAStereo\cite{cheng2020hierarchical} & 1.29 & 2.65& 1.51\\
    CREStereo\cite{li2022practical} &	             1.33 &	             \textbf{2.60} &	   1.54 \\
    Ours &	          1.34 &	              3.03 &	   1.62 \\
    \bottomrule
  \end{tabular}
  \caption{Benchmark results on KITTI 2015 test sets. Our approach is  close to the current best model performance. \textbf{Bold} is best.}
  \label{tab6}
\end{table}
\begin{figure*}[htbp]
	\begin{minipage}{1\linewidth}
	    \centerline{\includegraphics[width=1\textwidth]{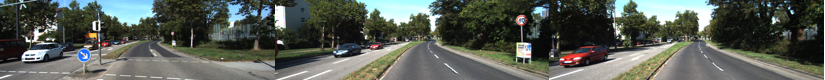}}
	    \vspace{-1pt}
	    \centerline{(a) Input Left Image}
		\vspace{5pt}
		\centerline{\includegraphics[width=1\textwidth]{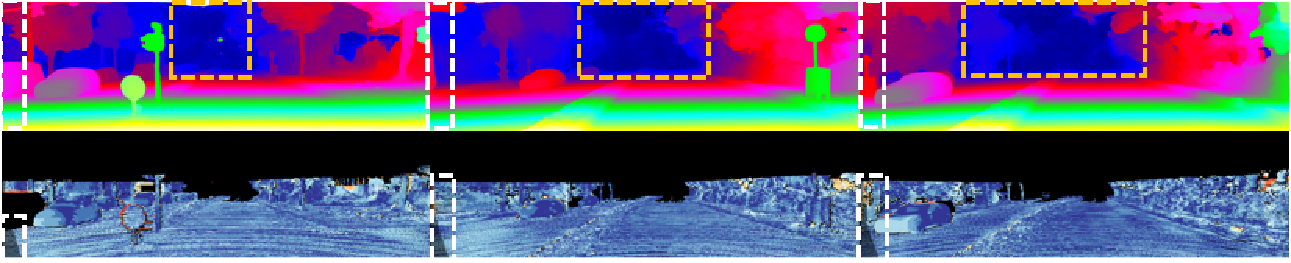}}
		\vspace{-2pt}
		\centerline{(b) ACVNet \cite{xu2022attention}}
		\vspace{5pt}
		\centerline{\includegraphics[width=1\textwidth]{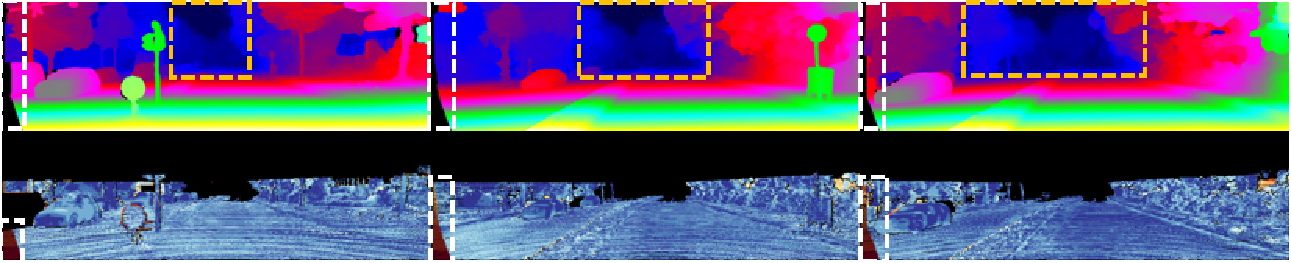}}
		\centerline{(c) Ours}
	\end{minipage}
	\caption{Visual presentation of KITTI 2015 test results. (a) From left to right, three consecutive test input left images of the same scene. (b-c) Detailed results map returned from KITTI 2015 online evaluation. From top to bottom, ACVNet and our model, respectively. The first row of each subfigure is the disparity map and the second row is the error map. Dark regions in the error maps denote the occluded pixels which fall outside the image boundaries. Our model has two special parts, one is the estimate the occluded boundary, and the other is the full disparity estimation in the distance region. We have specially annotated in the figures for distinction.}
	\label{fig4}
\end{figure*} 
\subsection{KITTI Benchmark Result}
For KITTI, we used all the images from KITTI 2012 and KITTI 2015 for fine-tuning, containing a total of 395 pairs of stereo images, we randomly selected 20 images for validation, and the rest were used for training. We trained a total of 400 epochs, the first 200 epochs with attention stride of s=3 and the last 200 epochs with attention stride of s=2. The test results of KITTI 2015 benchmark is shown in Table \ref{tab6}. AAUformer is comparable to state-of-the-art levels models. In addition, our method is able to identify the occluded regions which map to points outside the image domain in the other view, as shown in the error map of Fig \ref{fig4} and within the left boundary box marked, you can see that we could successfully estimate the occluded boundary distinct from other networks. Regrettably, the official KITTI website did not provide quantitative data metrics to evaluate occluded regions and sets disparity over 192 as occluded regions. Although this is not friendly to our network, we are still able to approach the top of the KITTI 2015 benchmark list. And more details on the definition of the Occlusion mask are in Appendix \textcolor{red}{B}.
\section{Conclusion}
Accurately recovering depth information from real scenes has always been a challenge for stereo matching, where lighting, occlusion and complex backgrounds are unavoidable obstacles. In this work, we proposed AAUformer, a full disparity stereo matching network based on the Transformer architecture that achieves comparable performance with
the state-of-the-art in several datasets and presents strong cross-generalization performance. We adopt several effective strategies to boost the matching accuracy of the model, mainly including windowed self-attention and multi-scale alternated attention backbone structure. The windowed self-attention delivers abundant local semantic features before the cross-matching of pixels. The multi-scale alternated attention backbone structure filters out invariant features to enhance the robustness of the model against natural environmental interference. In future work, thanks to our separated feature extraction and correlation network structure, we can try to introduce some pre-training weights of image semantic segmentation for better performance in real-world.

%%%%%%%%% REFERENCES
{\small

\bibliographystyle{plain}
\bibliography{egbib}
}

\clearpage
\section*{Appendix}
\appendix
\section{Loss Function}
\label{loss}
The final loss function is a combination of several different losses with different weights , including Relative Response loss ${L_{rr}}$ , smooth ${L_{1}}$ and binary-entropy loss ${L_{be}}$. The purpose of the ${L_{rr}}$ is to maximize the probability that the assignment matrix is close to 1 at the groundtruth, both for pixels that can be matched and for pixels that are considered to be occluded. ${L_{rr}}$ is denoted as
\begin{equation}
    {L_{rr}} = \frac{1}{{{N_R}}}\sum\limits_{i \in R} { - \log \left( {{t_i}} \right)}  + \frac{1}{{{N_Q}}}\sum\limits_{i \in Q} { - \log \left( {{t_{i,\phi }}} \right)}
\end{equation}
where the set $R$ represents the matched pixel region, the set $Q$ represents the occluded region, ${{t_i}}$ is the probability that the pixel matches the groundtruth, and ${t_{i,\phi}}$ is the probability that the pixel is occluded. Smooth ${L_{1}}$ is used to evaluate the difference between the groundtruth and the predicted disparity, applied to the raw disparity of low resolution, and the final disparity, respectively. 
${L_{1}}$ is denoted as
\begin{equation}
    {L_{d1,r}} = Smoot{h_{{L_1}}}\left( {{d_{gt}},{d_{raw}}} \right)
\end{equation}
\begin{equation}
    {L_{d1,f}} = Smoot{h_{{L_1}}}\left( {{d_{gt}},{d_{final}}} \right)
\end{equation}
where, ${d_{gt}}$ represents the groundtruth, ${d_{raw}}$ represents the raw predicted disparity at low resolution, and ${d_{final}}$ represents the predicted disparity of the final output after the context adjustment layer. ${L_{be}}$ is used to perform a binary classification for the final occlusion mask, can be denoted as 
\begin{equation}
    {L_{be,f}} = Entropy({y_i},{t_{i,\Phi }})
\end{equation}
where ${y_i}$ represents the category of the pixel, $y=1$ for the occlusion region, $y=0$ for the matching pixel region. In total, the final overall loss function is
\begin{equation}
    L = {w_1}{L_{rr}} + {w_2}{L_{d1,r}} + {w_3}{L_{d1,f}} + {w_4}{L_{be,f}}
\end{equation}
where $w$ represents the weight, which is set to 1 in all experiments.
\section{Occlusion of KITTI 2015}
\label{Occlusion}
\begin{figure}[htbp]
	\centering
	\begin{subfigure}{0.45\linewidth}
		\centering
		\includegraphics[width=1\linewidth]{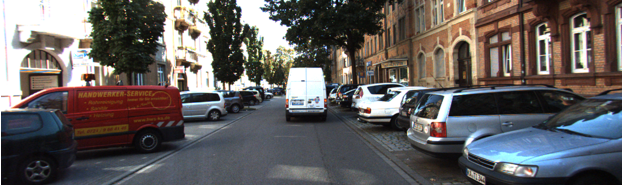}
		\caption{Left Image}
		\label{6-1}%文中引用该图片代号
	\end{subfigure}
	%\qquad
	%让图片换行，
	\begin{subfigure}{0.45\linewidth}
		\centering
		\includegraphics[width=1\linewidth]{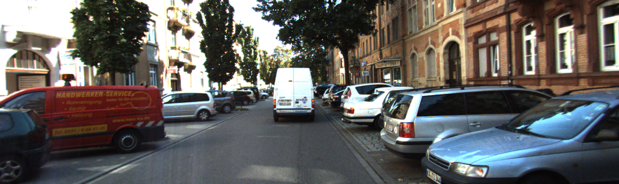}
		\caption{Right Image}
		\label{6-2}%文中引用该图片代号
	\end{subfigure}
	\begin{subfigure}{0.45\linewidth}
		\centering
		\includegraphics[width=1\linewidth]{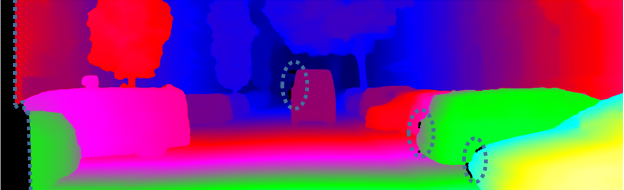}
		\caption{Ours}
		\label{6-3}%文中引用该图片代号
	\end{subfigure}
	\begin{subfigure}{0.45\linewidth}
		\centering
		\includegraphics[width=1\linewidth]{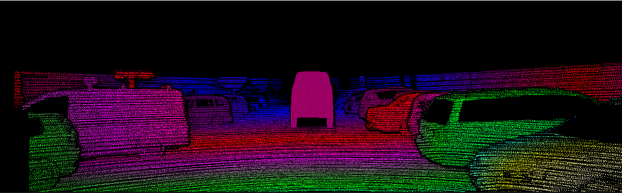}
		\caption{Groundtruth}
		\label{6-4}%文中引用该图片代号
	\end{subfigure}
	\caption{Visualization of a set of stereo images, including groundtruth and predicted disparity. (a-b) stereo image. (c) The result of the inference of our network. (d) The groundtruth are provided by the official. The dark area represents the occlusion mask }
	\label{Fig6}
\end{figure}
As shown in Fig. \ref{Fig6}, our predicted occlusion mask differs significantly from the mask presented by groundtruth, but this is interpretable. From Fig. \ref{6-4} we can see that the oclusion mask presented by KITTI 2015 is wide-ranging, including some untextured objects, far-away areas, and areas beyond the common field of view. However, the mask of our network prediction only needs to focus on these regions beyond the common field of view, because each of our disparity is obtained by calculating a unique corresponding pixel. Specifically, each pixel on the left can finds a corresponding pixel on the right with the maximum probability , and this maximum probability can be regarded as the confidence level. When the confidence of a pixel is lower than a certain value, the pixel can be considered unreliable and identified as a mask, and the opposite is matched. With this distinction, we generate disparity that are all veridical and reliable, which reduces the reliance on prior knowledge of the dataset.
\section{Computational Complexity Analysis}
\label{Complexity}
We all know that the Transformer model means more GPU memory, and the longer the sequence of input images, the more memory is consumed. Therefore, we use attention stride in all training process to ensure that there is no OOM (out of memory) during the training process. In this section, we compare our model with a similar Transformer model STTR. We use the torchstat library to calculate the parameters and flops of the model, while setting different attention stride to compare the GPU memory necessary for the inference process. From Table \ref{tab7}, the results show that our model has lower flops, time and memory consumption than STTR with the same parameter settings. The increase in the number of model parameters means our model is more robust, which can also be illustrated by the results of generalization experiments in Table \ref{tab4}.
\begin{table}
  \centering
  \begin{tabular}{c c c c c c}
    \toprule
    Model& Params &Stride & Flops &Time &Memory\\
    \midrule
    \multirow{3}{*}{STTR} & \multirow{3}{*}{6.6M} & 2 & 1850G & 1.92s & 15.1G \\
    ~ & ~ & 3 & 1146G & 0.75s & 6.5G\\
     ~ & ~ & 4 & 906G & 0.38s & 4.3G\\
    \multirow{3}{*}{Ours} & \multirow{3}{*}{10.8M} & 2 & 225G & 0.54s & 12.9G \\
    ~ & ~ & 3 & 148G & 0.21s & 5.3G\\
     ~ & ~ & 4 & 128G & 0.13s & 3.3G\\
     
    \bottomrule
  \end{tabular}
  \caption{The resolution of the input image is 960*540 and the device used for inference is an NVIDIA RTX3090 GPU with 24GB memory. All metrics are calculated using existing functions in the $torchstat$ library.}
  \label{tab7}
\end{table}

\section{Comparison For Different Alignment Levels}
\label{Alignment}
\begin{figure}[htbp]
    \centering
    \begin{subfigure}{0.45\linewidth}
        \centering
        \includegraphics[width=0.95\textwidth]{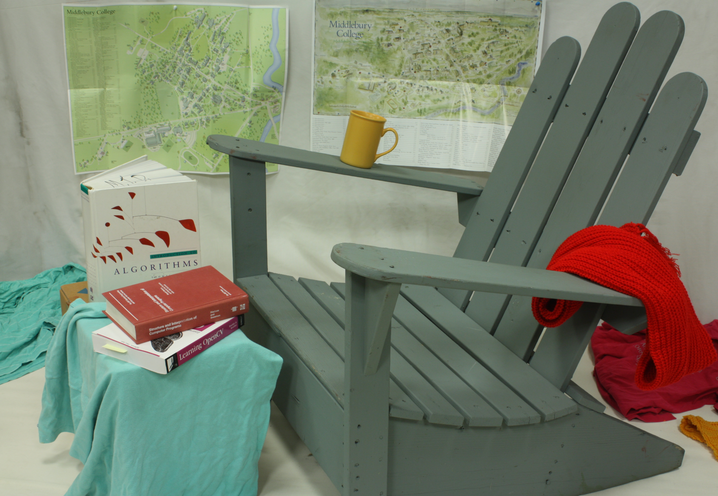}
        \caption{left}
    \end{subfigure}
    \centering
    \begin{subfigure}{0.45\linewidth}
        \centering
        \includegraphics[width=0.95\textwidth]{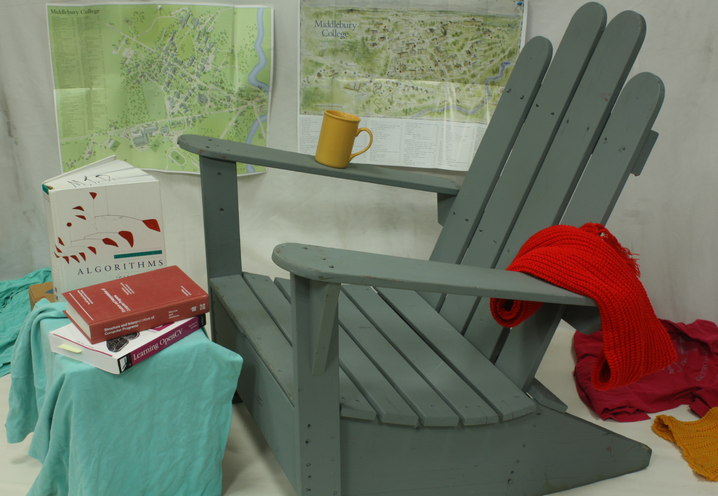}
        \caption{right}
    \end{subfigure}
    \centering
    \begin{subfigure}{0.45\linewidth}
        \centering
        \includegraphics[width=0.95\linewidth]{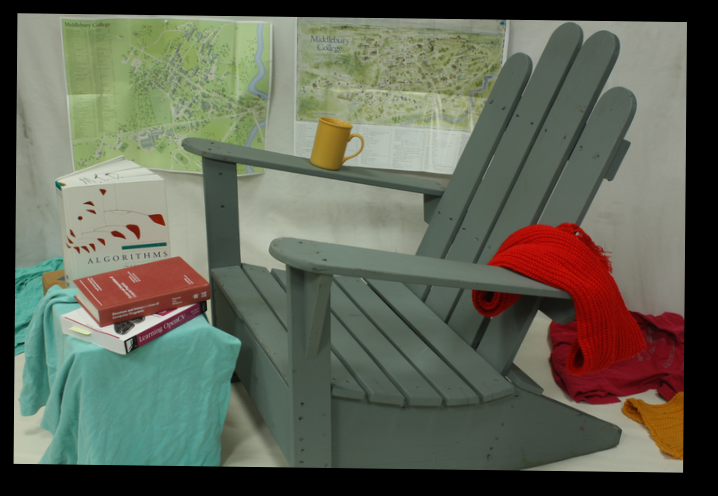}
        \caption{right+affine1}
    \end{subfigure}
    \centering
    \begin{subfigure}{0.45\linewidth}
        \centering
        \includegraphics[width=0.95\linewidth]{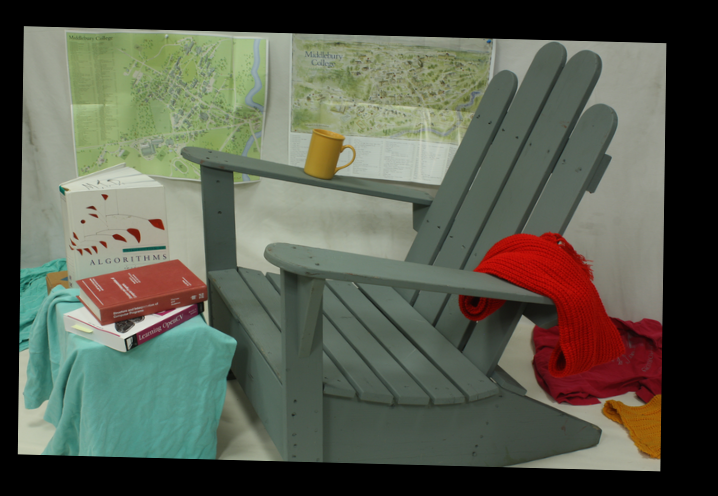}
        \caption{right+affine2}
    \end{subfigure}
    \centering
    \begin{subfigure}{0.45\linewidth}
        \centering
        \includegraphics[width=0.95\linewidth]{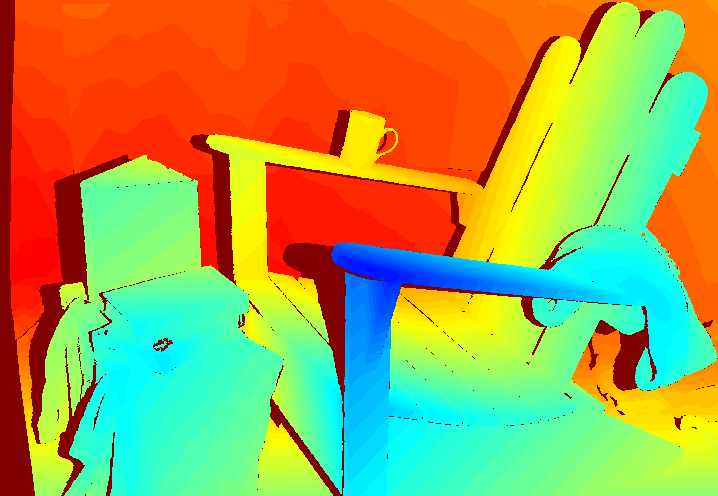}
        \caption{ground truth}
    \end{subfigure}
    \centering
    \begin{subfigure}{0.45\linewidth}
        \centering
        \includegraphics[width=0.95\linewidth]{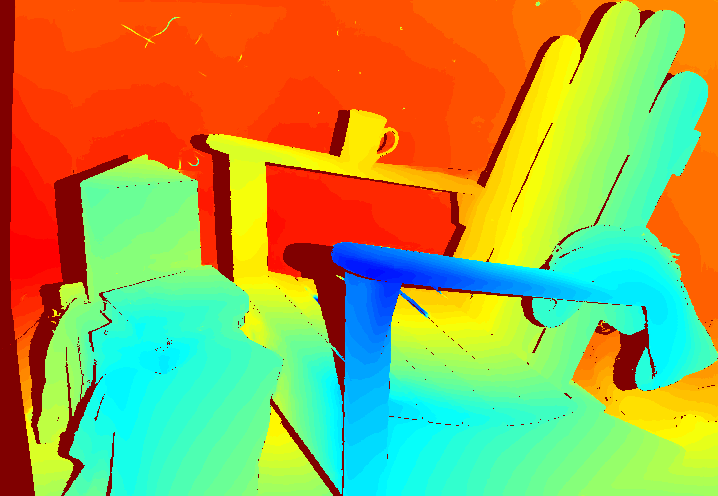}
        \caption{perfect alignment}
    \end{subfigure}
    \centering
    \begin{subfigure}{0.45\linewidth}
        \centering
        \includegraphics[width=0.95\linewidth]{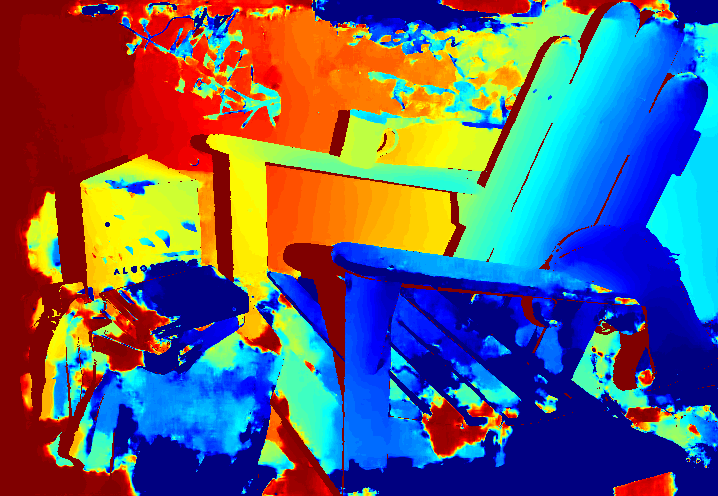}
        \caption{affine1}
    \end{subfigure}
    \centering
    \begin{subfigure}{0.45\linewidth}
        \centering
        \includegraphics[width=0.95\linewidth]{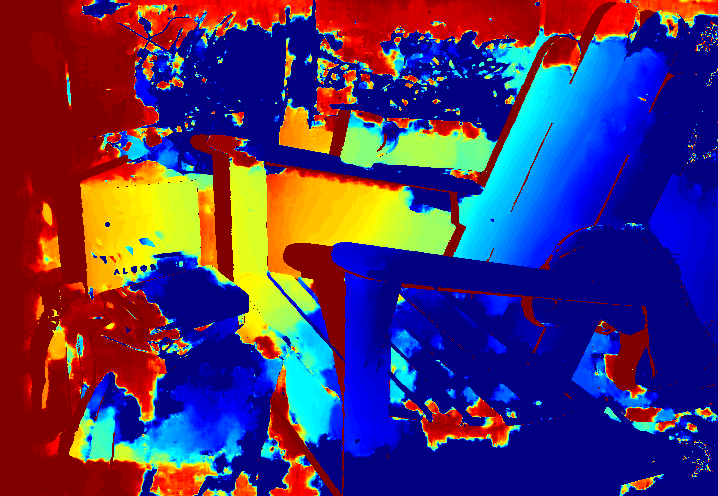}
        \caption{affine2}
    \end{subfigure}
    \caption{(a-b) A pair of stereo matched images after Rectified. (c-d) Right images after applying different degrees of affine transformation. (e-h) Disparity maps for different degrees of alignment }
    \label{affine}
\end{figure}

In this section, we will perform a simple experiment to show the effect of alignment level. We take a sample from Middlebury 2014 and apply different levels of affine transformation to it to simulate the alignment level. We used different right images for inference tests and got threshold pixel errors of $0.04($\%$)$, $22.2($\%$)$, and $26.2($\%$)$, respectively. We visualize the inference results for three different degrees of alignment in Fig \ref{affine}. As can be seen, the level of alignment affects the correct matching rate to some extent. Fortunately, in stereo vision, we are able to avoid such a case by using the external parameters of the camera for perfect alignment, and it is treated in most practical applications.
\section{More visualization results for PCA}
\begin{figure}[htbp]
    \centering
    \begin{subfigure}{0.45\linewidth}
        \centering
        \includegraphics[width=0.95\textwidth]{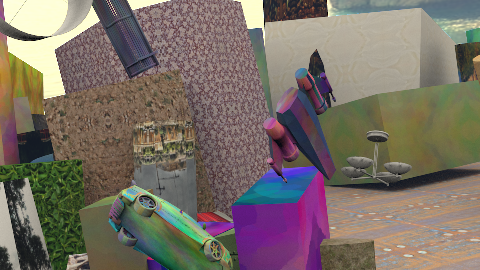}
        \caption{left}
    \end{subfigure}
    \centering
    \begin{subfigure}{0.45\linewidth}
        \centering
        \includegraphics[width=0.95\textwidth]{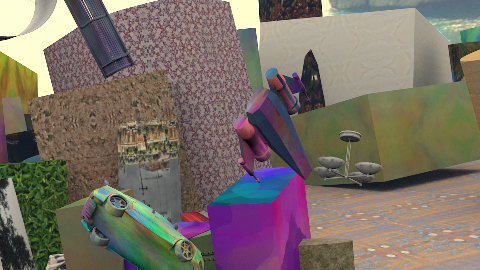}
        \caption{right}
    \end{subfigure}
    \centering
    \begin{subfigure}{0.45\linewidth}
        \centering
        \includegraphics[width=0.95\linewidth]{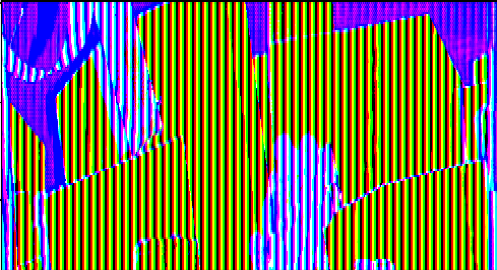}
        \caption{left feature}
    \end{subfigure}
    \centering
    \begin{subfigure}{0.45\linewidth}
        \centering
        \includegraphics[width=0.95\linewidth]{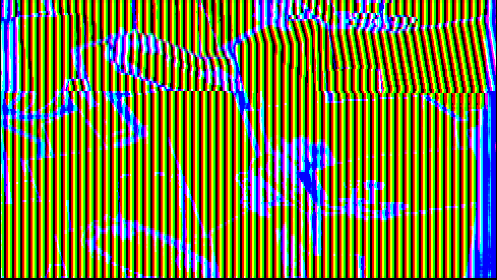}
        \caption{right feature}
    \end{subfigure}
    \centering
    \caption{(a-b) A pair of stereo matched images from Scene Flow. (c-d) visualization results for feature map.}
    \label{PCA_ALL}
\end{figure}

We visualize the left and right feature maps of the last layer through mapping the number of channels to 3. As shown in Fig \ref{PCA_ALL}, it can be noticed from the visualization results that the overall feature map also approximates a cyclic variation.
\label{pca_result}
\end{document}